\documentclass[a4paper]{article}
\usepackage{microtype}
\usepackage{graphicx}
\usepackage{subfigure}
\usepackage{algorithm}
\usepackage{algorithmic}
\usepackage{natbib}
\usepackage{booktabs} 
\usepackage{hyperref}

\usepackage{amsmath}
\usepackage{amssymb}
\usepackage{mathtools}
\usepackage{amsthm}
\usepackage[capitalize,noabbrev]{cleveref}

\theoremstyle{plain}
\newtheorem{theorem}{Theorem}[section]
\newtheorem{proposition}[theorem]{Proposition}

\theoremstyle{definition}

\theoremstyle{remark}

\makeatletter
\def\lstAZ{A, B, C, D, E, F, G, H, I, J, K, L, M, N, O, P, Q, R, S, T, U, V, W, X, Y, Z}
\def\lstaz{a, b, c, d, e, f, g, h, i, j, k, l, m, n, o, p, q, r, s, t, u, v, w, x, y, z}

\def\lstAZBB{B, C, D, E, F, G, H, I, J, K, L, M, N, O, P, Q, R, T, U, V, W, X, Y, Z}

\newcommand{\MkScr}[1]{\expandafter\def\csname s#1\endcsname{\mathscr{#1}}}
\newcommand{\MkUp}[1]{\expandafter\def\csname u#1\endcsname{\mathrm{#1}}}
\newcommand{\MkFrak}[1]{\expandafter\def\csname f#1\endcsname{\mathfrak{#1}}}
\newcommand{\MkCal}[1]{\expandafter\def\csname c#1\endcsname{\mathcal{#1}}}
\newcommand{\MkBB}[1]{\expandafter\def\csname #1#1\endcsname{\mathbb{#1}}}

\@for\i:=\lstAZ\do{%
	\expandafter\MkScr \i  %
	\expandafter\MkFrak \i  %
	\expandafter\MkUp \i %
	\expandafter\MkCal \i  %
		  }    
\@for\i:=\lstaz\do{%
	\expandafter\MkUp \i   }    
	
\@for\i:=\lstAZBB\do{%
	\expandafter\MkBB \i     }
	    
\makeatother

\newcommand{\dd}{ \mathrm{d}}

\DeclareMathOperator*{\argmin}{argmin}

\renewcommand{\epsilon}{\varepsilon}

\renewcommand{\tilde}{\widetilde}

\usepackage[textsize=tiny]{todonotes}

\begin{document}

\title{\vspace*{-5em} Stochastic Gradient Piecewise Deterministic Monte Carlo Samplers}
\author{Paul Fearnhead \\
  School of Mathematical Sciences, \\
  Lancaster University\\
  \texttt{p.fearnhead@lancaster.ac.uk} \\
  \and 
  Sebastiano Grazzi \\
  Department of Statistics,\\ University of Warwick \\
\texttt{sebastiano.grazzi@warwick.ac.uk} \\
  \and
  Chris Nemeth \\
   School of Mathematical Sciences,\\
  Lancaster University\\
  \texttt{c.nemeth@lancaster.ac.uk} \\
  \and
  Gareth Roberts\\
  Department of Statistics,\\ University of Warwick \\
  \texttt{gareth.o.roberts@warwick.ac.uk}
  }
\date{}
\maketitle

\begin{abstract}
Recent work has suggested using Monte Carlo methods based on piecewise deterministic Markov processes (PDMPs) to sample from target distributions of interest. PDMPs are non-reversible continuous-time processes endowed with momentum, and hence can mix better than standard reversible MCMC samplers. Furthermore, they can incorporate exact sub-sampling schemes which only require access to a single (randomly selected) data point at each iteration, yet without introducing bias to the algorithm's stationary distribution. However, the range of models for which PDMPs can be used, particularly with sub-sampling, is limited. We propose approximate simulation of PDMPs with sub-sampling for scalable sampling from posterior distributions. The approximation takes the form of an Euler approximation to the true PDMP dynamics, and involves using an estimate of the gradient of the log-posterior based on a data sub-sample. We thus call this class of algorithms stochastic-gradient PDMPs. Importantly, the trajectories of stochastic-gradient PDMPs are continuous and can leverage recent ideas for sampling from measures with continuous and atomic components. We show these methods are easy to implement, present results on their approximation error and demonstrate numerically that this class of algorithms has similar efficiency to, but is more robust than, stochastic gradient Langevin dynamics.
\end{abstract}

{\bf{Keywords}}: Bouncy Particle Sampler; Control-variates; MCMC; Stochastic gradient Langevin dynamics; Sub-sampling; Zig-Zag Sampler

\section{Introduction}\label{sec: intro}
Whilst Markov chain Monte Carlo (MCMC) has been the workhorse of Bayesian statistics for the past thirty years, it is known to scale poorly for large datasets, since
each iteration of MCMC requires the evaluation of the log-posterior density which
typically scales linearly with data size.
As a result, approximate MCMC methods that use only a subsample of data at each iteration have become popular. The first such method, and arguably the most widely used, is the stochastic gradient Langevin dynamics (SGLD) algorithm of \citet{welling2011bayesian}. The idea of SGLD is to approximately simulate a Langevin diffusion that has the posterior as its stationary distribution. The method involves two approximations: firstly it simulates an Euler discretisation of the Langevin diffusion; and secondly it approximates the gradient of the log-posterior, i.e. the drift of the diffusion, based on a subsample of the data. 
The SGLD algorithm has been applied to applications such as topic models \cite[]{baker2018large}, Bayesian neural networks \cite[]{gawlikowski2021survey} and probabilistic matrix factorisation \cite[]{csimcsekli2017parallelized}; and has also been extended to more general dynamics - e.g. \citet{chen2014stochastic, ma2015complete}. See \citet{nemeth2021stochastic} for a review.

Recently there has been interest in developing efficient continuous time MCMC algorithms 
known as Piecewise Deterministic Markov processes (PDMP) \cite[]{vanetti2017piecewisedeterministic, davis1984piecewise}. Such methods include the Bouncy Particle Sampler \cite[]{bouchard2018bouncy} and the Zig-Zag algorithm \cite[]{bierkens2019} amongst others \cite[]{fearnhead2018}. Due to their non-reversibility, these methods often mix better than traditional reversible MCMC algorithms \cite[]{diaconis2000analysis,bierkens2016non}. Furthermore, \citet{bierkens2019} show how these methods can be implemented whilst only using a small subsample of data at each iteration, and yet still target the true posterior distribution. PDMPs (without subsampling) have been also developed and successfully applied in computational physics for example for the simulation of hard sphere models \cite[]{Peters_2012, PhysRev, 10.3389/fphy.2021.663457}.

However PDMP samplers can be challenging to implement, particularly when using subsamples, as they require bounding the gradient of the log-density. There has been some work on automating the simulation of PDMPs using numerical methods \cite[]{corbella2022automatic,pagani2020nuzz}, but these methods largely require exact calculation of the gradient of the log-posterior at each iteration and are incompatible with subsampling ideas.   

In this paper, we investigate the approximate simulation of PDMPs with subsampling as a means of achieving scalable MCMC. The idea is to discretise time into intervals of length $\epsilon$, for each interval we sample a single data point, and then simulate the (approximate) dynamics of the PDMP using only the information from this data point. Loosely speaking, this can be viewed as simulating an Euler approximation to the dynamics of the PDMP with subsampling. This simple idea was suggested, but not investigated, by \citet{bertazzi2022approximations} and \cite{bertazzi2023splitting}. We call these methods stochastic-gradient PDMPs, or SG-PDMPs. We show that they give results that are competitive with SGLD, but have the advantage of being more robust to large discretisation sizes. They simulate  continuous trajectories, and we implement a version that leverages this property to easily sample from a trans-dimensional posterior distribution for models that incorporate variable selection.

\begin{figure}
    \centering
    \includegraphics[width = 0.45\textwidth]{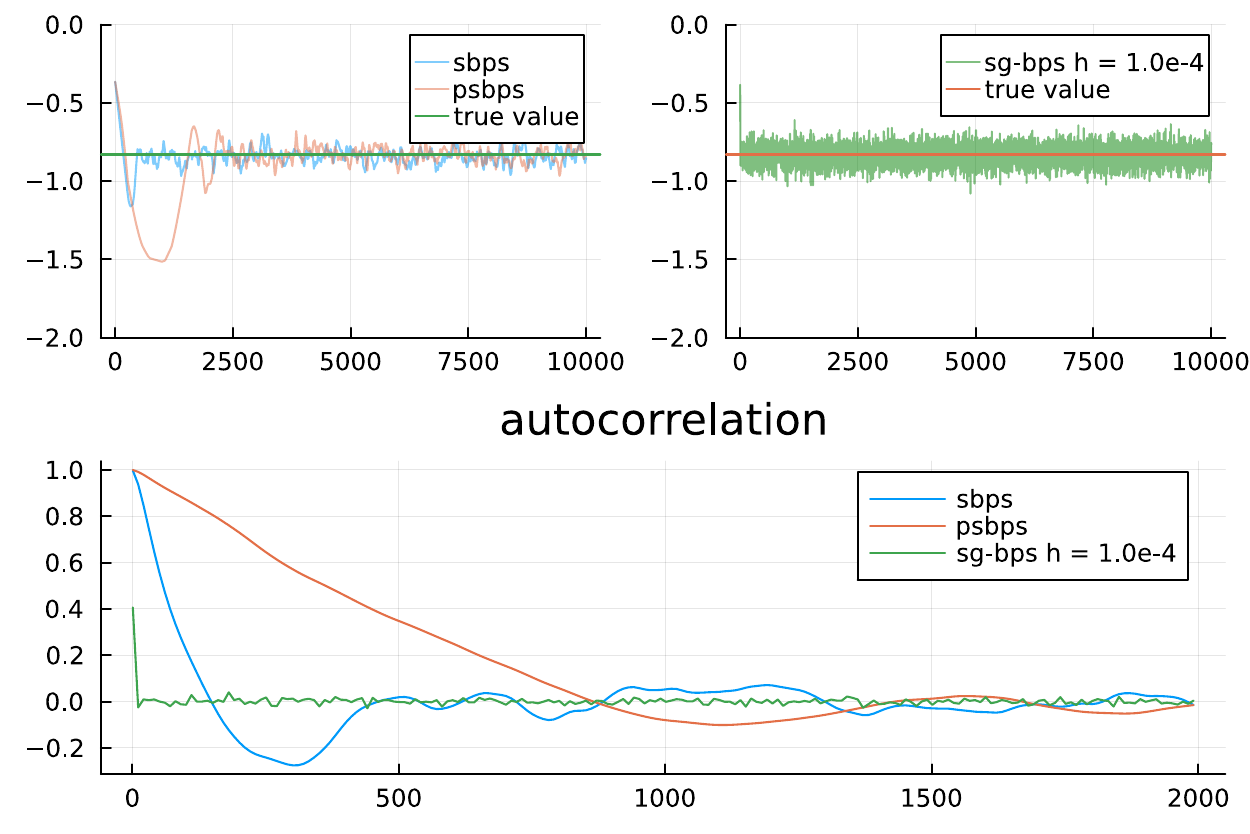}
    \caption{Top panels: traces of SBPS, SBPS with pre-conditioning (PSBPS) (left) and our Euler approximation of the bouncy particle sampler, SG-BPS (right) with step-size $h = 10^{-4}$. Bottom panel: auto-correlation function of the first coordinate for the three algorithms. All algorithms were implemented with the same CPU cost, and output thinned to give 10,000 samples.}
    \label{fig: summary_comparison}
\end{figure}

The idea of approximately simulating a PDMP algorithm with subsampling has been previously suggested by \citet{pakman2017stochastic}. 
Their algorithm, called the stochastic bouncy particle sampler (SBPS) uses Poisson thinning to simulate events of the Bouncy Particle Sampler with subsampling. The idea is that, if we can upper bound the actual event rate, and can simulate events with this upper bound rate, then we can thin (or remove) events with an appropriate probability to obtain events simulated with the required rate. For most target distributions we cannot calculate an upper bound for the rate, so \citet{pakman2017stochastic} suggests estimating this upper bound based on information on the rate from sub-samples of data at proposed event times. In practice one of the main differences between our proposal and SBPS is that, for each discrete time-step we only need to sample a single data point. By comparison the stochastic bouncy particle sampler uses sub-sample batch sizes $n$ that are of the order of, say 10\% of, the full-data size $N$ (a smaller sub-sample batch size would compromise its performance). Empirical results suggest our ability to use a sub-sample of one can lead to a substantial increase in efficiency. For example Figure~\ref{fig: summary_comparison} shows the trace of the first coordinate and the autocorrelation function of two versions of the SBPS algorithm and of our SG-BPS algorithm,  for a Bayesian logistic regression problem (see Appendix~\ref{app: comparison SBPS} for details).  In Appendix~\ref{app: comparison SBPS}, we present a more detailed comparison of our SG-PDMPS and the SBPS, for a logistic regression model and a linear regression model. In all cases we found SBPS mixes more slowly for a fixed computational cost. Also SBPS sometimes does not converge if started in the tail of the posterior. Our ability to use, at every iteration, a mini-batch size of 1 data point is a key advantage of our method and, in most contexts, we expect worse performance (relative to CPU cost) for larger mini-batches (see Appendix~\ref{sec: Comparison of SG-PDMPs different batch-sizes} for numerical simulations varying the batch-sizes of SG-PDMPs).


The paper is structured as follows. Section~\ref{subsec: pdmps with subsampling} reviews PDMPs with subsampling, while in Section~\ref{sec: stochastic gradient PDMP samplers} we describe in detail SG-PDMPs and present the main theoretical results which shows the order of error for this class of algorithms. Section~\ref{sec: linear regression} highlights the main benefit of using SG-PDMPs on an illustrative linear regression example. In Section~\ref{sec: numerical experiments}, our algorithms are applied to a logistic regression model, with artificial data, and for a Bayesian neural network model with a variety of real datasets. Finally, Section~\ref{sec: discussion} outlines promising research directions stemming from this work.

\section{PDMP samplers} \label{subsec: pdmps with subsampling}

Throughout we will consider the problem of sampling from a target density on $\RR^d$ of the form $\pi(x) \propto \exp(-U(x))$, and assume that
\begin{equation}
    \label{eq: energy function}
U(x) = \sum_{j=1}^N U_j(x)
\end{equation}
for some large value $N$ and differentiable functions $U_j:\RR^d \to \RR, \, j=1,2,\ldots, N$. This is a common problem in, for example, Bayesian inference where $\pi(x)$ is the posterior density and the factors, $U_j(x)$ for $j=1,\ldots,N$, corresponds to either the log prior or the log-likelihood of conditionally independent observations.

Like other MCMC algorithms, a PDMP sampler simulates a Markov process that is designed to have $\pi(x)$ as its stationary distribution. However, whilst standard MCMC algorithms simulate discrete-time, and generally reversible, Markov process, a PDMP sampler simulates a continuous-time, non-reversible process.

The dynamics of the PDMP are defined in the product space of position and velocity $E = \RR^d \times \cV$ where $\cV \subset \RR^d$. Evolution of the process is Markovian on $E$ and given (in our cases) by piecewise constant velocity dynamics interspersed by a sequence of random event times at which
the velocity component of the PDMP changes. 

We will denote the state of the PDMP by $z = (x,v) \in E$, with $x$ denoting its position and $v$, its velocity. Let $\phi_s(z)$ denote the change in state due to the deterministic dynamics over a time-period of length $s$. That is $z_{t+s}=\phi_s(z_t)$, where
\[
\phi_s(z_t)=\phi_s( (x_t,v_t) )= (x_t+sv_t,v_t).
\]

The dynamics of the PDMP are then specified through the rate at which events occur and how the velocity changes at each event. As a PDMP is Markovian, the instantaneous rate, $\lambda(z)$ of an event only depends on the current state.
The change of velocity at an event will be defined through a Markov kernel 
that only acts on the velocity. 

There are many choices of event rate and Markov kernel that will have $\pi(x)$ as the $x$-marginal of the stationary distribution of the resulting PDMP. We are interested in PDMP samplers that use sub-sampling ideas, for which the rate and transition kernel can each be written as a sum of terms, each of which depends on just a single factor $U_j(x)$. We call these S-PDMPs.

\subsection{PDMP samplers with subsampling}\label{sec: example s-pdmps}

The idea of the S-PDMP is that we define the dynamics in terms of dynamics associated with each factor $U_j(x)$. For each factor, $j=1,\ldots,N$, introduce a reflection $F^{j}_x(\cdot)$, such that if $v'=F^{j}_x(v)$ then $v=F^{j}_x(v')$. The rate of events in the S-PDMP is then of the form
\[
\lambda(z) = \frac{1}{N}\sum_{j=1}^N \lambda^j(z)
\]
where $\lambda^j\colon E \to \RR^+$ satisfies
\[
\lambda^j(x, F_x^{j}(v)) - \lambda^j(x, v) =  N \left(v\cdot \nabla U_j(x) \right) 
\] and the Markov kernel for the transition at an event allows for $N$ possible transitions. If the state immediately before the event is $z=(x,v)$, then the state after the event is $z'=(x,v')$ with
\[
\Pr(v'=F^{j}_x(v))= \frac{\lambda^j(z)}{N\lambda(z)}, \mbox{ for $j=1,\ldots,N.$}
\]
It can be shown that for these choices, the S-PDMP targets $\pi(x)$ \cite[]{bierkens2019,fearnhead2018,chevallier2021pdmp}.


The $N\nabla U_j(x)$ term that appears in the rate $\lambda^j(z)$ can be viewed as an unbiased estimator of $\nabla U(x)$. Just as control variates can be used to reduce the variance of such an estimator, we can transform each factor $U_j(x)$ to $\hat{U}_j(x)$ so that $N\nabla \hat{U}_j(x)$ is a better approximation of $\nabla U(x)$. This is done by defining $\hat{U}_j(x)=U_j(x)-x(U_j(\hat{x})-\sum_{j=1}^N U_j(\hat{x}))$, for some centering value $\hat{x}$. This gives
\begin{equation}
\label{eq: stochatic gradeint control variates}
\nabla \hat{U}_j(x)= N\left\{\nabla U_j(x)-\nabla U_j(\hat{x})\right\} +\sum_{j=1}^N \nabla U_j(\hat{x}).
\end{equation}
We will use such a set of transformed factors in the following, with $\hat{x}$ assumed to be an estimate of the mode of $\pi$, obtained through an initial run of stochastic gradient descent. 
However, the following arguments will apply to other choices. As we will show, and as is consistent with SGLD \cite{baker2018large}, choices of factors that give lower variance estimators of $\nabla U(x)$ will lead to better algorithms. 

\subsubsection{Bouncy Particle Sampler with subsampling}\label{subsec: bps}

The original BPS \citep{bouchard2018bouncy} takes values in $\RR^d\times\RR^d$ and targets a density proportional to $\pi(x)p(v)$, where $p(v)$ is a density for the velocity that is independent of position and is symmetric.  There are two common choices for $p(v)$, one is a uniform density over a $d$-dimensional hypersphere, and the other is a standard $d$-dimensional Gaussian distribution. More generally, we could consider any distribution defined in terms of an arbitrary distribution for the speed $||v||$, together with an independent uniform distribution for the direction of $v$, i.e. $v/||v||$. The dynamics of the BPS are unaffected by this choice, other than the initialisation, which should involve drawing $v_0$ from $p(v)$, and at the refresh events (see below).

When using subsampling, the reflection event associated with factor $j$ is 
\[
F^j_x(v)=v - \frac{  v\cdot \nabla \hat{U}_j(x) 
}{\|\nabla \hat{U}_j(x)\|^2}\nabla \hat{U}_j(x),
\]
and the rates are
\begin{equation}
    \label{eq: poisson rate}
    \lambda^j(z) = \max( v\cdot \nabla \hat U_j(x) 
    , 0).
\end{equation}
Furthermore, to ensure the process is irreducible, with constant rate $\lambda_{\mathrm{ref}}>0$, the velocity component is refreshed with an independent draw $v \sim p(v)$. 

\subsubsection{Zig-Zag sampler with subsampling}\label{subsec: zz}
The Zig-Zag sampler with subsampling \citep{bierkens2019} has velocity $v \in \{-1,1\}^d$. It involves $d$ possible events, each of which flips one component of the velocity. 

As there are $d$ possible events, an additional superscript, $i$, is introduced for each type of flip. For $i=1,\ldots,d$ we define 
\[
\mathrm{R}_{i}(v):=(v_1,v_2,\dots,v_{i-1}, -v_i, v_{i+1},\dots,v_d),
\]
i.e. only the $i$th component is flipped. This transition will be the same for all factors, $j$, that is $F_x^{i,j}(v)=\mathrm{R}_{i}(v)$. The rate associated with this transition is 
\begin{equation} \label{eq:ratezz}
\lambda^{i,j}(z) = \max( v \partial_{x_i} \hat{U}_j(x), 0),
\end{equation}
where, by analogy to eq.~\eqref{eq: stochatic gradeint control variates},
\[
\partial_{x_i} \hat{U}_j(x) = N\left\{\partial_{x_i} U_j(x)-\partial_{x_i} U_j(\hat{x})\right\} +\sum_{j=1}^N \partial_{x_i} U_j(\hat{x}).
\]

\subsection{Simulating S-PDMPs}

S-PDMPs are continuous-time Markov processes that have $\pi(x)$ as the $x$-marginal of their stationary distribution. 
For reasons of practicality, we require a method for simulating these processes. The challenge in simulating an S-PDMP lies in simulating the event-times, as the other dynamics are simple. 

Given current state $z=(x,v)$,
the rate of the next event just depends on the time until the next event, which we define as $\lambda_z(s) := \lambda((x+vs,v))$, which uses the fact that, if $z_t=z$ and there has been no further event, then at time $t+s$ the state is $z_{t+s}=(x+vs,v)$. Thus the time until the next event will be the time of the first event in an in-homogeneous Poisson process (IPP) of rate $\lambda_z(s)$. We will denote the time until the next event as $\tau \sim \text{IPP}(s \to \lambda_z(s))$. If this rate is constant, $\tau$ is distributed as an exponential random variable with rate $\lambda(z)$, i.e. $\tau \sim \text{Exp}(\lambda(z))$.

There exists a range of techniques for simulating from an in-homogeneous Poisson process. The most common general methods are based on the idea of thinning \cite[]{lewis1979simulation}: we upper bound the rate $\lambda(\phi_s(z))$ by a piecewise linear function of $s$; we simulate possible events with this upper bound rate; and we accept these as actual events with probability equal to the true rate divided by the bound. The challenge with efficiently sampling an S-PDMP then comes from finding good upper bounds for the rates.

\citet{bierkens2019} show that if we obtain an upper bound, $\lambda^+_z(s)$, that bounds the rate corresponding to each factor of the target distribution, which in other words bounds $\lambda^j(\phi_s(z))$ for $j=1\ldots,N$, then we can simulate the S-PDMP exactly whilst accessing only a single factor at each iteration. This works by (i) simulating the next possible event, at a rate $\lambda^+_z(s)$; (ii) sample a factor uniformly at random; (iii) if the possible event is at time $\tau$, and we have sampled the $j$th factor, we accept the event with probability $\lambda^j(\phi_\tau(z))/\lambda^+_z(\tau)$, and change the state according to the kernel $Q_j(\phi_\tau(z),\cdot)$. 
Importantly, given the upper bound, the only step that depends on the target is step (iii) and this involves only a single factor. 

In settings where S-PDMPs can be applied, they are shown to have an overall complexity of $\cO(1)$, relative to the sample size $N$. This is theoretically motivated and numerically illustrated for logistic regression models in \citet{bierkens2019} and \citet{bierkens2020boomerang}. However, the set of models where we can find appropriate upper bounds is currently limited and motivates our interest in approximate stochastic gradient PDMP samplers. 

\section{Stochastic gradient PDMP samplers}\label{sec: stochastic gradient PDMP samplers}

By analogy with stochastic gradient Langevin algorithms, where one approximately simulates a Langevin diffusion with $\pi(x)$ as its stationary distribution, we introduce stochastic gradient PDMP algorithms that approximately simulates an S-PDMP algorithm, and called SG-PDMP algorithms. The idea of simulating approximations to PDMPs is introduced in \citet{bertazzi2022approximations, bertazzi2023splitting}. 
 
Following \citet{bertazzi2022approximations}, our approximation is based on discretising time into intervals of size $\epsilon$, and for each interval choosing a factor at random and then simulating the dynamics of the underlying S-PDMP based on the rate and dynamics for that factor. As a starting point, and to further simplify the simulation of any event in the time interval, we fix the event rate based on the state at the start of the time interval, and simulate at most one event. As we show below, using these two simplifications does not impact the order of the approximation. 

The resulting general algorithm is given in Algorithm \ref{alg:sg-pdmp}. We assume our true PDMP has $K$ types of events, and introduce a rate and transition for each type and each factor. For example, with the Zig-Zag sampler, we have $d$ event types, one for each possible component of the velocity to flip. See eq.~\eqref{eq:ratezz} for the rates associated with each pair of event type and factor, with event $i$ having the same transition regardless of the factor, and with $v'=R_i(v)$ for the new velocity. For the Bouncy Particle Sampler, we have two types of events. The first is a reflection, with rate for factor $j$ given by \eqref{eq: poisson rate}, and transition given by $v'=F_x^j(v)$. The second type of event is a refresh event. This is the same for all factors and has constant rate $\lambda_{\mathrm{ref}}$  with transition $v$ drawn from the stationary distribution for the velocity.

The algorithm then loops over time intervals of length $\epsilon$, simulates a factor $J$, and then an event time for each type of event. These occur with a constant rate defined as the rate for each event $i$, and factor $J$, for the current state of the process. We then calculate the time, $\tau$, and event type $i^*$ which occurs first. If that event occurs within the interval, we simulate the exact continuous-time dynamics of the PDMP over the time interval with an event of type $i^*$ for factor $J$ at time $\tau$. If the first event does not occur within the time interval, we just simulate the PDMP dynamics over the time interval with no events. 


\begin{algorithm}[hbt!]
\caption{Stochastic gradient PDMP sampler}\label{alg:sg-pdmp}
\begin{algorithmic}
\REQUIRE $(x,v) \in E$, step size $\epsilon > 0$, time horizon $T>0$. 
\STATE $t = 0$, 
\STATE Rates $\lambda^{i,j}(z)$ for event of type $i$ associated with potential $U_j$, for $i=1,\ldots,K$ and $j=1,\ldots,N$. 
\STATE Transition kernel $Q^{i,j}(\cdot;z)$ associated with event of type $i$ and potential $U_j$.
\WHILE{$t < T$}
\STATE $J \sim \mathrm{Unif}(1,2,\dots,N)$
\STATE $\tau_i \sim \mathrm{Exp}( 
\lambda^{i,J}(z)
),\mbox{ for } i=1,\ldots,K$ 

\STATE $\tau=\min_{i=1,\ldots,K}\,\tau_i, \, i^* =\argmin \tau_i$
\IF{$\tau < \epsilon$}
\STATE  $x \gets x + v\tau$
\STATE $v \sim Q^{i^*,J}(\cdot;(x,v))$  \COMMENT{Simulate new velocity}
\STATE $x \gets z +v(\epsilon-\tau)$
\ELSE
    \STATE $x \gets x + v\epsilon$
\ENDIF
    \STATE $t \gets t + \epsilon$
    \STATE Save $(x,t)$
\ENDWHILE
\end{algorithmic}
\end{algorithm}

Using results in \citet{bertazzi2022approximations} we can show that the resulting SG-PDMP sampler can give an $O(\epsilon)$ approximation to the distribution of the true PDMP over any fixed time interval $t$:
\begin{proposition} \label{prop:1}
Let $\overline{\mathcal{P}}_t(z,\cdot)$ and  ${\mathcal{P}}_t(z,\cdot)$ be the transition kernels for the stochastic gradient PDMP process of Algorithm \ref{alg:sg-pdmp} and for the true underlying PDMP process, respectively. Assume the PDMP processes have bounded velocities, so $\|v\|<C_0$ for some $C_0$. Assume that for any state $z=(x,v)$ and any $i=1,\ldots,K$ and $j=1,\ldots,N$ the function $\lambda^{i,j}(x+vt,v)$ has a continuous derivative with respect to $t$. Then there exists constants $C(z,T)$, that depend on the initial state, $z$ and time interval $T$, and $\epsilon_0>0$ such that for all $\epsilon<\epsilon_0$
\[
||{\mathcal{P}}_T(z,\cdot)-\overline{\mathcal{P}}_T(z,\cdot)||_{TV} \leq C(z,T)\epsilon.
\]
\end{proposition}
See Appendix \ref{sec:proof} for the proof.

While this result shows that the SG-PDMP algorithm is an $O(\epsilon)$ approximation of the true S-PDMP, a more informal analysis gives insight into the approximation error. Consider the probability of simulating an event in the next interval of length $\epsilon$ given a current state $z=(x,v)$. Let $\lambda^j(z)=\sum_{i=1}^K \lambda^{i,j}(z)$, then the probability for the true S-PDMP of no event in an interval of length $\epsilon$ is $\exp\{-\sum_{j=1}^N (1/N)\int_0^\epsilon \lambda^j(x+vt,v)\mbox{d}t\}$, whereas for the SG-PDMP it is $(1/N)\sum_{j=1}^N \exp\{-\lambda^j(z)\epsilon\}$. We can decompose the difference into two terms
\begin{eqnarray*}
\lefteqn{\exp\left\{-\sum_{j=1}^N \int_0^\epsilon \frac{\lambda^j(x+vt,v)}{N}\mbox{d}t\right\} - \exp\left\{-\sum_{j=1}^N \frac{\lambda^j(z)\epsilon}{N}\right\}}\\  &+ &\exp\left\{-\sum_{j=1}^N \frac{\lambda^j(z)\epsilon}{N}\right\} - \frac{1}{N}\sum_{j=1}^N \exp\left\{ -\lambda^j(z)\epsilon\right\}.
\end{eqnarray*}
The first difference is due to the use of constant event rates over the interval. Assuming the rates are Lipschitz continuous, the change in event rate over an interval of length $\epsilon$ is $O(\epsilon)$, so the difference in the integral of the rates is $O(\epsilon^2)$. We could reduce this error by using a better approximation to the event rate over the interval \cite[]{bertazzi2023splitting}. The second difference is due to the stochastic gradient approximation, that is we consider just one factor for each time interval. There are two points to make. The first is that by Jensen's inequality, this difference is always negative. This means that the stochastic gradient approximation reduces the probability of an event. As the PDMPs only introduce events when moving into areas of lower probability density, the impact of this is that it samples from a heavier-tailed approximation to the target. Second, we can use a Taylor expansion to get the highest order, in $\epsilon$, term of the approximation. This is the $O(\epsilon^2)$ term, as the first order terms cancel, whose coefficient is
\[
- \left( \frac{1}{N}\sum_{j=1}^N \lambda^j(z)^2 - \left(\sum_{j=1}^N \frac{\lambda^j(z)}{N}\right)^2 \right),
\]
which is (minus) the variance of the $\lambda^j(z)$s. This means that the approximation error depends on the variability of the estimator of the gradient of $U(x)$ across the different factors. Furthermore, this error is $O(\epsilon^2)$ which means that using a better approximation for how the rates vary over the time interval would not improve the rate in $\epsilon$ of the approximation we introduce. Proposition~\ref{prop:1} provides an order of approximation in terms of the transition kernels of our algorithms. For results on the approximation of the invariant measure, we refer to Section 6 in \citet{huggins2017quantifying}: under appropriate mixing conditions, the order of the error (in terms of step-size) of the approximation in the transition density will imply a similar order of error in the invariant measure.

One disadvantage of Algorithm~\ref{alg:sg-pdmp} for coarse discretisations, i.e. when $\epsilon$ is large, is that it does not allow more than one event in the interval. Thus a simple improvement is, if there is an event at time $\tau<\epsilon$ within the interval, to then iterate the simulation of events for the time interval $[\tau,\epsilon]$. See Appendix~\ref{sec:sgzzbps} for details of the resulting algorithm for stochastic gradient versions of the Zig-Zag sampler and the Bouncy Particle Sampler. These are the algorithms we evaluate in the empirical results in Sections \ref{sec: linear regression} and \ref{sec: numerical experiments}.


\subsection{Extension with sticky components for variable selection}\label{sec: sg-pdmp sticky} 
The methods developed here can be naturally extended to the PDMPs with boundary conditions recently developed in \citet{chevallier2020reversible, bierkens2021sticky, chevallier2021pdmp, bierkens2023methods} for sampling from target densities which are only piecewise smooth and for reference measures which are mixture of continuous and atomic components.  The only difference these methods have is with the dynamics between events. For example, if there is a boundary then the PDMP may reflect off the boundary if it hits it. 

For sampling from Bayesian posteriors for models with variable selection, where the target distribution is for the co-efficients of the model, it is common to have a prior that has positive probability on co-efficients being zero. The sticky PDMP of \citet{bierkens2021sticky} can deal with this by setting the velocity of components of $x$ to zero if that component of $x$ is zero, and then re-introducing the non-zero velocity at a certain rate. This is also easily incorporated into the SG-PDMP algorithm by appropriately changing the dynamics of the process.
See Appendix \ref{sec:szz} for further details and the corresponding SG-PDMP algorithms.




\begin{figure*}[!htbp]
\centering
\begin{minipage}{0.3\textwidth}
\includegraphics[width=\textwidth]{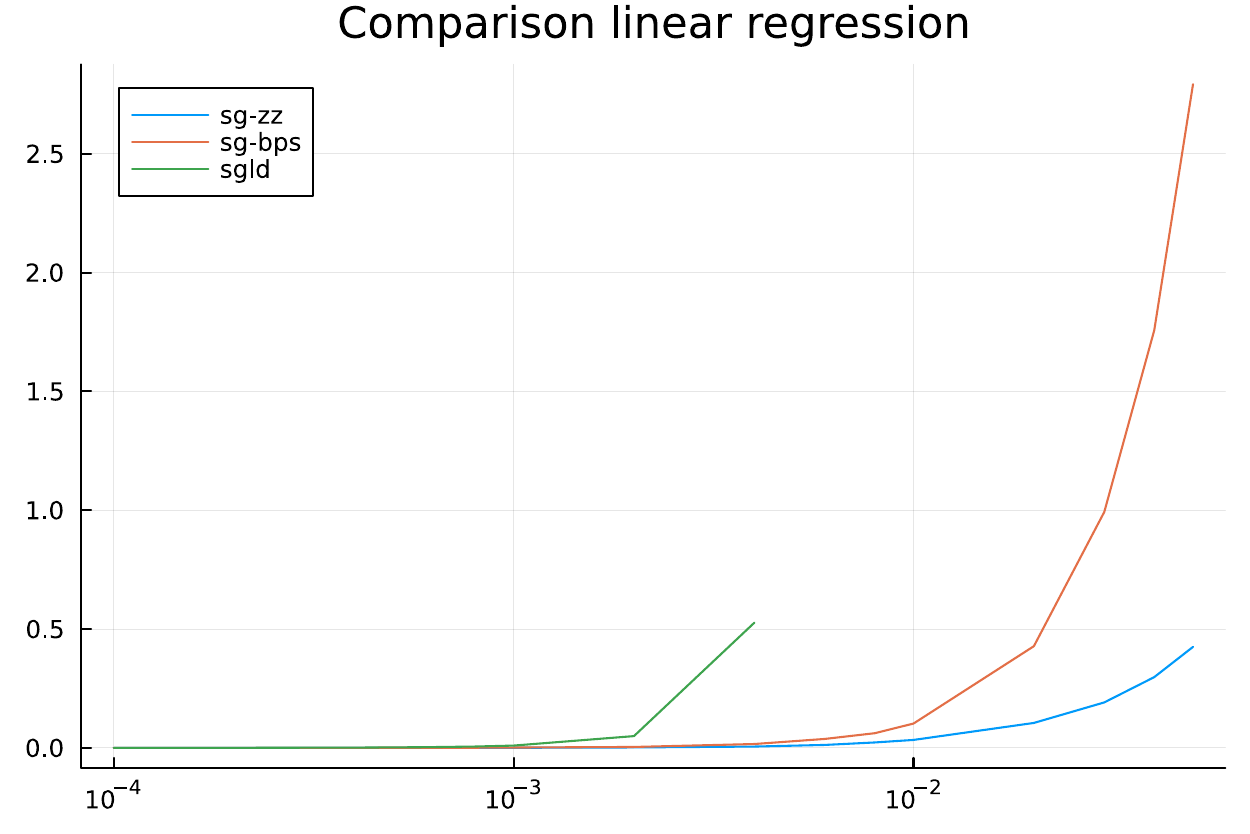} \\
\includegraphics[width=\textwidth]{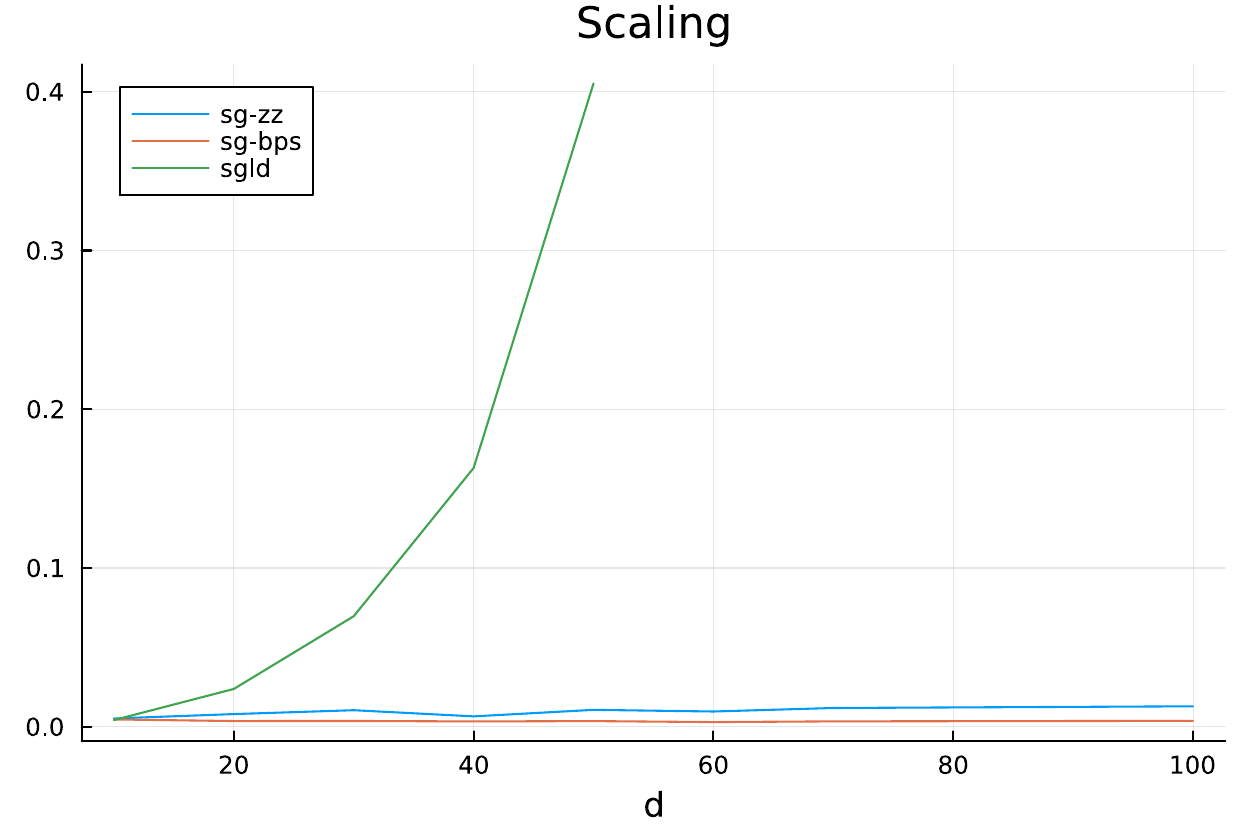}
\end{minipage}
\begin{minipage}{0.65\textwidth}
\includegraphics[width=\textwidth]{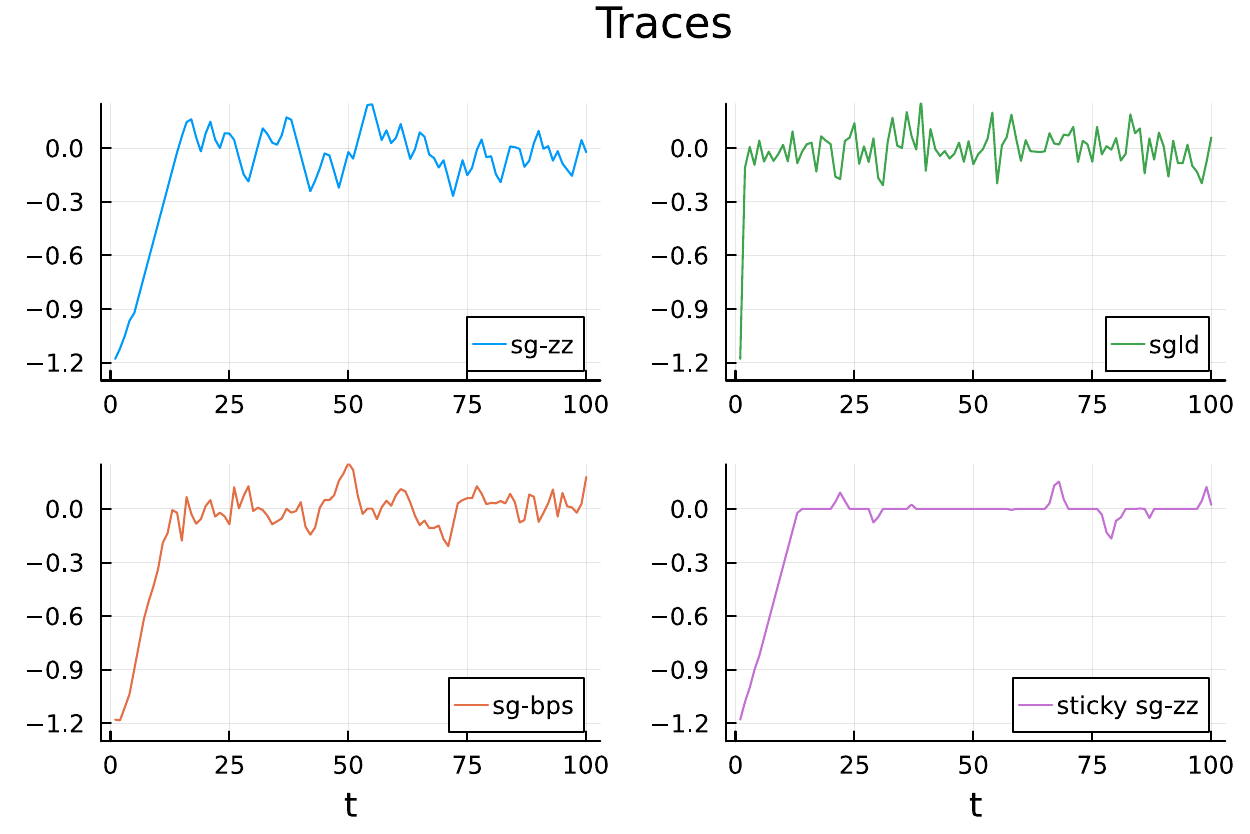}
\end{minipage}
\caption{\label{fig: linear_regression}
  Left top panel: error in standard deviation estimation of each coordinate \eqref{eq: error on variances} $\cE^{(h,d)}$ as a function of the stepsize $h$ ($x$-axis in log-scale). Left bottom panel: $\cE^{(h,d)}$ as a function of $d$. Right panels: trace plots of the first coordinate. }
\end{figure*}
\section{Illustrative example: Linear regression model}\label{sec: linear regression}

We illustrate the  benefits of SG-PDMP samplers on a simple linear regression model 
$y = Ax + \epsilon$, for a response variable $y \in \RR^N$, covariates $A \in \RR^{N \times d}$,  parameters $x\in \RR^d$ and $\epsilon \sim \cN(0,N c \mathrm{I}_N)$ (the variance of the noise is re-scaled by a factor to aid comparison across different values of $N$, as the posteriors will have similar variance regardless of $N$). For this problem, we compare sample averages computed with the trajectories of each sampler against the true expectation of selected functionals of the true posterior. 

 We set $A_{:,1} = \mathbf{1}$ to account for the intercept and simulate the data as $A_{i,j} \sim \cN(0,\Sigma)$ with $\Sigma_{i,j} = \Sigma_{j,i} \sim \text{Unif}(0.4, 0.8)^{|i-j|}, i = 1,2,\dots,d; \,j = 2,3,\dots,d$ and set $x_1 = 0$. We simulate $x_i \overset{\mathrm{iid}}{\sim} \cN(0,1), i = 2,3,\dots,d$ with prior $\cN(0,100\times \mathrm{I}_d)$ for $x$.
 
  We consider two experimental settings. 
First, we simulate $N = 10^6$ covariates with $d=5$ features and run SGLD, SG-ZZ and SG-BPS for different step-sizes and $T = 5\times 10^6$ iterations, initialising each sampler 
at the ordinary least squares estimate. Figure~\ref{fig: linear_regression} (left top panel) shows 
\begin{equation}\label{eq: error on variances}
        h \mapsto \cE^{(d,h)} = \frac{1}{d}\sum_{i=1}^d \left(\frac{\hat \sigma_{i}^{(h)} - \sigma_i}{\sigma_i}\right)^2
    \end{equation}
for each sampler,  where $\hat \sigma^{(h)}_i$ is the sample standard deviation of the $i$th component estimated with the trace. The $x$-axis is displayed on a log-scale. We also consider the setting where we initialise the algorithms in the tails of the distribution and plot in Figure~\ref{fig: linear_regression} (left panels) the first few iteration of the trace of the first coordinate (whose true value was set to 0) and including the SG-SZZ sampler.
    
Second, we set $N = 10^5$ and simulate multiple datasets with varying $d = 10,\dots,10^2$. For each dataset, we run SGLD, SG-ZZ and SG-BPS. We fix the step size equal to $h = 5\times10^{-7}$ and the number of iterations $T = 2\times 10^6$, such that all samplers have a comparable error for the dataset with the smallest number of dimensions. Figure~\ref{fig: linear_regression} (bottom left panel) shows $d \mapsto \cE^{(d,h)}$ for each sampler.           
In all simulations, each sampler uses a stochastic gradient with control variates and 1 data point at each iteration \eqref{eq: stochatic gradeint control variates}. The number of gradient evaluations and the running time is comparable among the different algorithms.

 This simple tractable example provides useful insight into the advantages of SG-PDMP  over SGLD.  Firstly we see greater stability of SG-PDMPS for large discretisation steps. This is in contrast with SGLD which diverges to infinity if the discretisation step is too large. (In these simulations, the maximum step size that can be taken for SGLD before diverging is 0.004.) Secondly, the error in SG-PDMPs grows much slower compared to SGLD as the dimensionality of the parameter increases: in these simulations, SGLD was unstable for $d > 50$. This offers opportunities for SG-PDMPs especially for high dimensional  problems. Finally, the deterministic dynamics of SG-PDMPs allows these algorithms to deal easily with the variable dimensional posterior in the case where we use a prior on each component on $x$ that includes a point-mass at zero. This is achieved using the stochastic gradient version of the sticky Zig-Zag sampler of Section \ref{sec: sg-pdmp sticky}. The trace of the sampler is shown in the bottom-right panel, and we see it successfully enforces sparsity of the first parameter, whose true value is equal to 0. 

\section{Numerical experiments}\label{sec: numerical experiments}
In this section, we  assess the performance of SG-PDMPs and SGLD on intractable posterior distributions. We consider 
logistic regression and a Bayesian neural network model.

We analyse the performance 
of each sampler by computing various metrics to assess bias and numerical accuracy. In the logistic regression example, we simulate a large dataset
so that the posterior distribution is well-approximated by a Gaussian distribution formed from a Laplace approximation. Then, for each sampler, we compute the sum of squared errors  as in eq.~\eqref{eq: error on variances}  between the estimated standard deviations and the standard deviations of the Laplace approximation.  

The samplers considered here are approximate and asymptotically biased, which means that standard MCMC diagnostics such as Effective Sample Size (ESS) are not appropriate \citep{nemeth2021stochastic}. 
A popular performance metric for stochastic gradient Monte Carlo algorithms is the Kernel Stein Discrepancy \citep{gorham2017measuring}; see Appendix~\ref{app: Stein discrepancy kernel}. This metric can detect both poor mixing and bias from the target distribution. 

For all the problems considered, predictive performance of each sampler was assessed by splitting the dataset $\cD$ into a training set $\cT$ and its complementary test set $\cT^c$ so that $\cT \sqcup \cT^c =  \cD$. We set $|\cT| = 0.9 |\cD|$. Each datum $\cD_i = (X_i, y_i)$ consists of a pair of covariates together with a dependent variable. Then, for each point $x$ of the output of each algorithm, its predictive accuracy is given by \begin{equation}
    \label{eq: loss function}
    \frac{1}{|\cT^c|}\sum_{(X_i,y_i) \in \cT^c } \ell(X_i, y_i, x)
\end{equation}
where $\ell$ is a non-negative loss function between $y_i$ and the predicted outcome given the covariates $X_i$ and parameter $x$. Specific loss functions will be specified in context.


Each algorithm was implemented $M$ times, each time with a different random permutation of training and test datasets. Average loss was computed across these $M$ realisations.
In all examples, control-variate  stochastic gradient estimates were employed
(Equation~\eqref{eq: stochatic gradeint control variates}) and samplers were initialised
at the same control variate point, with no burn-in period. The control variate was computed using the stochastic optimization algorithm ADAM \citep{kingma2014adam}, with $10^6$ iterations and a subsample size equal to 1\% of the dataset and all the other parameters suggested therein.

\subsection{Logistic regression}
A Bayesian logistic regression problem was considered
with standard Gaussian prior with variance $10$ for each component. We set the dimension of the parameter $p = 10$ and we simulate $N = 10^5$  covariates $X_{i,j} \sim \cN(0, \Sigma)$ where $\Sigma_{i,j} = \Sigma_{j,i} \sim \text{Unif}(-\rho, \rho)^{|i-j|}, \, \rho = 0.4$ for $i = 1,2,\dots,N$ and $j = 1,2,\dots,p$.  $y$ was simulated with true parameters: $x^\star_i \sim \cN(0,1)$.

\begin{figure}
    \centering
    \includegraphics[width = 0.45\textwidth]{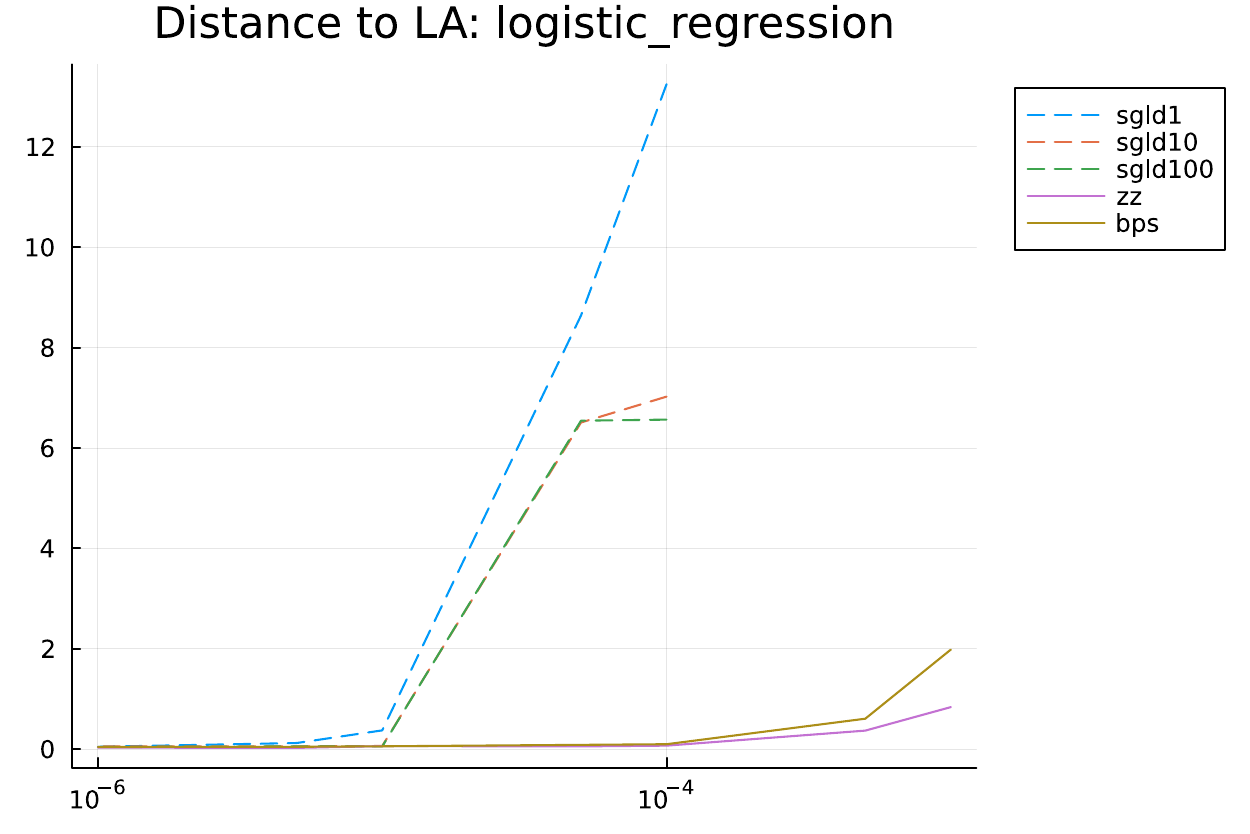}
    \caption{Error between standard deviation estimation of each coordinate and the one relative to the Laplace approximation as in \eqref{eq: error on variances} for the logistic regression as a function of the stepsize $h$ ($x$-axis in log-scale). Dashed lines corresponds to SGLD algorithms, solid lines to SG-PDMPs.}
    \label{fig: distance to Laplace approximation}
\end{figure}

SGLD was implemented with minibatches of sizes 1, 10 and 100 data points and compared against SG-PDMP samplers SG-ZZ, SG-BPS and SG-SZZ. All algorithms were run for $T = 10^6$ iterations. 
The running time, and number of gradient evaluations, for SGLD with minibatches of sizes 10 and 100 were significantly higher compared to the SG-PDMP algorithms.  
Each simulation was implemented for several values of $h = 10^{-6}, \dots, 10^{-3}$. Figure~\ref{fig: distance to Laplace approximation} shows $h \to \cE^{(h,d)}$ - eq.~\eqref{eq: error on variances} between the estimated standard deviation and the standard deviation of the Laplace approximation. In  Appendix~\ref{app: logistic}, we give the Kernel Stein discrepancy and the loss function \eqref{eq: loss function}. For all experiments, SG-PDMPs performs similarly to SGLD for small step-sizes and out-perform SGLD with 1, 10, 100 mini-batch sizes for larger step-sizes (the maximum step-size for SGLD before the algorithms diverge is $10^{-4}$).  

A second experiment 
considering an over-parameterised regime was also considered. In this setting we took $p = 10^2$ parameters and $N = 10^2$ covariates, with true parameter being 0 with probability 0.5. SGLD, SG-ZZ, SG-BPS, SG-SZZ were all implemented for $T = 10^7$ and $h = 10^{-4}$. SG-SZZ utilised a spike-and-slab prior with spike weight equal to $w = 0.5$. Table~\ref{tab: sparse logistic} shows the mean squared error between true parameters and sample mean and median.
\begin{table}[h]
\begin{centering}
\begin{tabular}{l|llll}
       & SGLD & SG-ZZ & SG-BPS & SG-SZZ \\\hline
mean  & 3.08155 & 3.15546  & 2.93989  & \textbf{1.42291}    \\
median & 3.02602  & 3.12318 & 2.90635  & \textbf{1.18081}
\end{tabular}
\caption{Mean squared error relative to the sample mean and sample median for over-parameterised logistic regression.} \label{tab: sparse logistic}
\end{centering}
\end{table}

\subsection{Bayesian neural networks}

Bayesian approaches to neural networks (BNNs)
provides a powerful calculus to quantify uncertainty and reduce the risk of overfitting
by incorporating techniques such as dropout \citep{gal2016dropout} and imposing sparsity-inducing priors \citep{polson2018posterior}. 

Three datasets were considered from the \emph{UCI machine learning repository}\footnote{https://archive.ics.uci.edu/}, varying in dimension and data size labelled as \textit{boston}, \textit{concrete}, \textit{kin8mn}. A two-layer Bayesian neural network was utilised:
\[
y = a_2(W_2(a_1(W_1x + b_1) + b_2) + N(0,1)
\]
where $a_1(x) = x$  and $a_2(x) = \max(0, x)$, for unknown parameters $W_1 \in \RR^{50 \times p}, W_2 \in \RR^{1 \times 50}, b_1 \in \RR^{50}$ and $b_2 \in \RR$ for $p$ covariates in each dataset. An independent prior $\cN(0, 10)$ was chosen for each component of $x = (W_1, W_2, b_1,b_2)$.  With this setting, the number of dimensions of the parameter $x$ ranged from 501 to 751. 

\begin{figure}
    \centering
    \includegraphics[width = 0.49\textwidth]{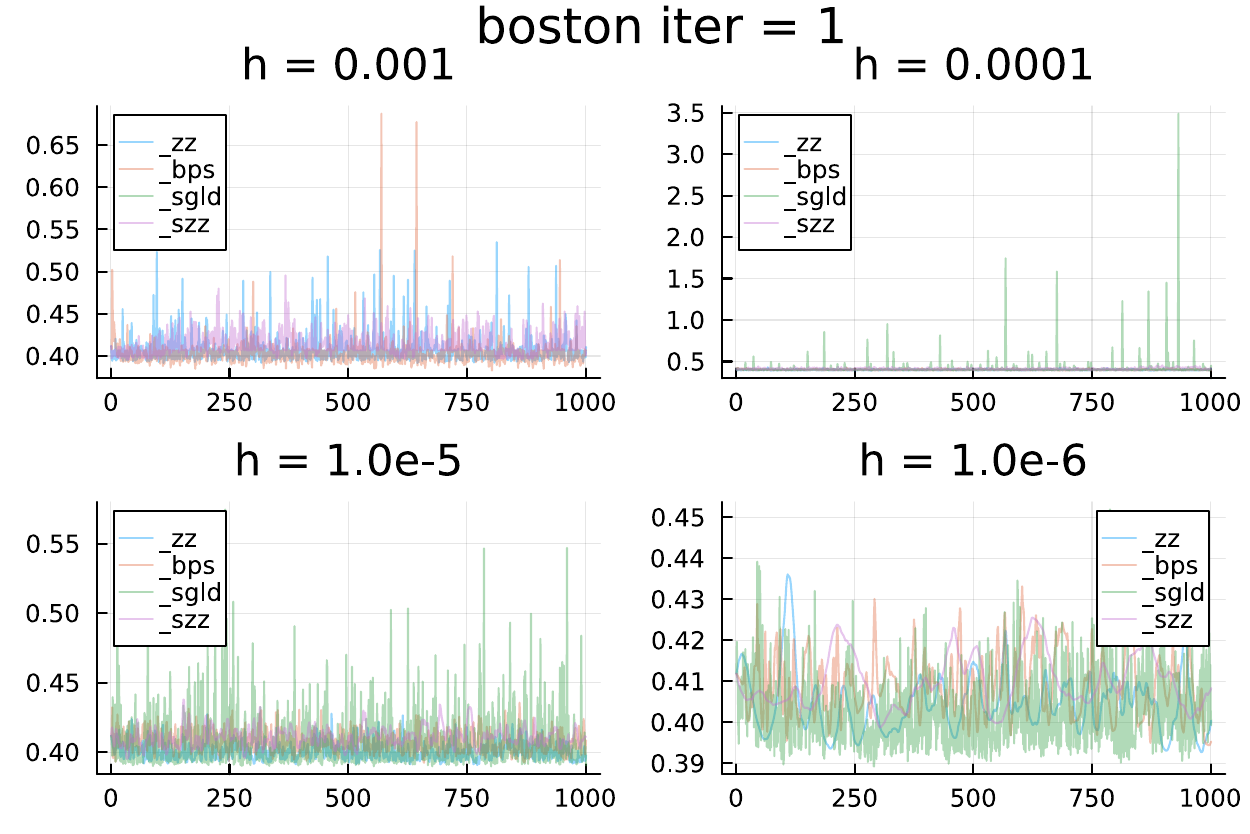}
    \caption{Trace of the loss function for each sampler for different step-sizes $h$ relative to the first permutation of the training and test set of the dataset `boston'.}
    \label{fig:loss nn}
\end{figure}

SG-ZZ, SG-SZZ, SG-BPS and SGLD were implemented for $N = 10^6$ iterations on $M=3$ random permutations of the training and test datasets, varying  $h = 10^{-6},\dots,10^{-3}$. The loss function was chosen to be the mean squared error. Figure~\ref{fig:loss nn} shows the loss function  trace plot for each sampler in the first permutation of training and test set, for the dataset `boston' which has $N = 507$ data points and $p = 13$ covariates. For this experiment, SGLD was unstable for $h = 10^{-3}$. In Table~\ref{tab: loss nn} we display the average loss of the trace which was computed by averaging the $M = 3$ random permutations of training and test datasets.
The results for the other three datasets considered here are qualitatively similar and can be found in Appendix~\ref{app: Bayesian neural networks}.

\begin{table}[h]
\label{tab: loss nn}
\begin{center}
\begin{tabular}{l|llll}\\
 & \multicolumn{4}{c}{\textbf{h}}\\
  \textbf{Sampler}      & $10^{-3}$ & $10^{-4}$ & $10^{-5}$ & $10^{-6}$ \\
        \hline
SGLD    &    \emph{NaN}        & \emph{0.4940} & \emph{0.4586} & \textbf{0.4526} \\
SG-ZZ   &    0.4720 & 0.4572  & \textbf{0.4518}  & 0.4555   \\
SG-BPS  &   \textbf{0.4692} & \textbf{0.4533} & 0.4543  & 0.4574 \\
SG-SZZ  & 0.4704 & 0.4586  & 0.4564 & \emph{0.4643}    
\end{tabular}
\end{center}
\caption{Average loss for each sampler for different values of step size $h$ relative to the dataset `boston'. In bold the best performance and italic the worst performance given a step size $h$.} 
\end{table}


\section{Discussion}\label{sec: discussion}
 We have presented SG-PDMPs as a competitive and robust alternative to the popular SGLD algorithm for approximate Bayesian inference. In particular, we highlighted that SG-PDMPs are generally more stable than SGLD for large step sizes, and can take advantage of the continuous-time sample paths through e.g. the sticky dynamics for regression with model choice. The better robustness that SG-PDMPs have is due to the fixed velocity dynamics, and is reminiscent of that of relativistic Hamiltonian dynamics \cite[]{pmlr-v54-lu17b}, and the stochastic gradient Barker dynamics of \cite{mauri2024robust}. 
 
 
 There are several natural extensions to this paper which could be explored as future work. It is known that suitable pre-conditioning can improve mixing of PDMPs. An adaptive preconditioning mass matrix for PDMPs has been studied in \citet{bertazzi2022adaptive} and \citet{pakman2017stochastic}. As noted in \citet{livingstone2022barker}, the convergence of adaptive algorithms can be drastically improved if algorithms are robust to the choice of step size, as SG-PDMPs appear to be. Second, SGLD is the Monte Carlo analogue of the popular  stochastic optimisation algorithm  \emph{Stochastic Gradient Decent} (SGD). Similarly, Piecewise Deterministic Markov processes which converge to global minima appeared for example in \citet{monmarche2014piecewise}. It is therefore natural to develop a stochastic optimization method based on SG-PDMP dynamics. Finally, it would be interesting to develop higher-order approximation schemes. This will be non-trivial, as, in Section~\ref{sec: stochastic gradient PDMP samplers}, we showed that higher order approximations of the Poisson rates, as developed in \citet{bertazzi2023splitting}, cannot be directly adopted in this context since the error given by taking the stochastic gradient dominates the overall error of the algorithm.


\bibliography{example_paper}
\bibliographystyle{royal}

\newpage
\appendix
\onecolumn


\section{Proof of Proposition \ref{prop:1}} \label{sec:proof}

The proof follows by application of Theorem 4.17 of \citet{bertazzi2022approximations}. Within \citet{bertazzi2022approximations}, the authors discuss approximations of the Zig-Zag sampler with sub-sampling -- see their Example 5.7. However they comment that such an approximation is different from the algorithms they consider and just sketch how their proofs could be extended to such an algorithm. Here we consider a different approach, by showing that we can re-formulate our SG-PDMP algorithm as an specific case of their Algorithm 3, and then directly apply a result for that algorithm.

Let the current state be $z$. The distribution of the time to the next, $\tau$, that is simulated in Algorithm \ref{alg:sg-pdmp} satisfies
\[
\Pr(\tau>t) = \sum_{j=1}^N \frac{1}{N} \exp\left\{
-t \sum_{i=1}^K \lambda^{i,j}(z)
\right\}.
\]
This follows by averaging  $\Pr(\tau>t|J=j)$ over the possible values of $j$.
Given an event at time $\tau$, type of event $i^*$ and factor $J$ has probability mass function
\begin{equation} \label{eq:algpmf}
\Pr(i^*=i,J=j|\tau) = \frac{\lambda^{i,j}(z)\exp\left\{-t\sum_{k=1}^K \lambda^{k,j}(z)\right\}}{\sum_{l=1}^{N} \left(\sum_{k=1}^K  \lambda^{k,l}(z) \right)\exp\left\{-t\sum_{k=1}^K \lambda^{k,l}(z)\right\}}.
\end{equation}
This follows as, by Bayes theorem, this is proportional to the density of choosing factor $j$, having an event at time $\tau$ and the event being of type $i$ -- which is the term in the numerator.  The denominator is then just the normalising constant of the probability mass function.

In Algorithm 3 of \citet{bertazzi2022approximations} they simulate the time $\tau$ such that
\begin{equation} \label{eq:b1}
\Pr(\tau>t) = \exp \left\{ -\int_0^t \bar{\lambda}(z,s) \mbox{d}s \right\},
\end{equation}
for some time-inhomogeneous rate $\bar{\lambda}(z,s)$, that is the sum of $m$ event specific rates $\bar{\lambda}(z,s) =\sum_{k=1}^m \bar{\lambda}_k(z,s)$. The the type of event, $k$ say, is simulated with probability $\bar{\lambda}_k(z,\tau)/\bar{\lambda}(z,\tau)$.

To relate the two algorithms, we first let $m=KN$. We then slightly adapt the notation of \citet{bertazzi2022approximations} and subscript the event rates by the pair $(i,j)$, rather than a single index. Define the event specific time-inhomogeneous rates by, for $i=1,\ldots,K$ and $j=1,\ldots,N$,
\begin{equation} \label{eq:rates}
\bar{\lambda}_{i,j}(z,s) = \lambda_{i,j}(z)
\frac{\exp\left\{-s\sum_{k=1}^K \lambda^{k,j}(z) \right\}}{\sum_{l=1}^{N} \exp\left\{-s \sum_{k=1}^K \lambda^{k,l}(z) \right\}} .
\end{equation}
Then 
\[
\bar{\lambda}(z,s)=\sum_{j=1}^N\sum_{i=1}^K \lambda^{i,j}(z) 
\frac{\exp\left\{-s\sum_{k=1}^K \lambda^{k,j}(z) \right\}}{\sum_{l=1}^{N} \exp\left\{-s \sum_{k=1}^K \lambda^{k,l}(z) \right\}} ,
\]
and by noting that
\[
\frac{\mbox{d}}{\mbox{d}s} \log \left\{\sum_{l=1}^{N} \frac{1}{N} \exp\left\{-s \sum_{k=1}^K \lambda^{k,l}(z) \right\} \right\} = -\bar{\lambda}(z,s),
\]
\[
\int_0^t \bar{\lambda}(z,s)\mbox{d}s = \left[-\log \left\{\frac{1}{N} \sum_{l=1}^{N} \exp\left\{-s \sum_{k=1}^K \lambda^{k,l}(z) \right\} \right\}\right]_0^t = 
-\log(1) + \log \left\{\frac{1}{N} \sum_{l=1}^{N} \exp\left\{-t \sum_{k=1}^K \lambda^{k,l}(z) \right\} \right\}.
\]
Substituting into eq.~\eqref{eq:b1}, we see that the distribution of $\tau$ in Algorithm 3.1 of \citet{bertazzi2022approximations} is the same of the distribution of $\tau$ simulated by Algorithm \ref{alg:sg-pdmp}. Conditional on $\tau$, the probability of choosing event $i,j$ in Algorithm of \citet{bertazzi2022approximations} is
\[
\frac{\bar{\lambda}_{i,j}(z,\tau)}{\bar{\lambda}(z,\tau)} = \frac{\lambda^{i,j}(z)\exp\left\{-s\sum_{k=1}^K \lambda^{k,j}(z) \right\} }{ \sum_{l=1}^N
\left(\sum_{k=1}^K \lambda^{k,l}(z)\right) \exp\left\{-s\sum_{k=1}^K \lambda^{k,l}(z) \right\}
},
\]
which is the same as the probability for Algorithm \ref{alg:sg-pdmp} -- see  (\ref{eq:algpmf}).

Thus Algorithm \ref{alg:sg-pdmp} is equivalent to Algorithm 3 of \citet{bertazzi2022approximations}, with the specific choice of rates, $\bar{\lambda}_{i,j}(z,s)$ given in (\ref{eq:rates}). Proposition \ref{prop:1} follows immediately from Theorem 4.17 of \citet{bertazzi2022approximations} if we can show their Assumption 4.14 holds. As they comment (see their Note 4.16), this will hold if the state of the true and approximate PDMP has bounded norm for a finite time horizon, and if their Assumption 4.6 holds. The former is true as we are considering PDMPs with bounded velocity. 

Their Assumption 4.6 requires (A) an $\bar{M}(z)$ such that for $0\leq s \leq \epsilon \leq \epsilon_0$, and all $i,j$
\[
|\bar{\lambda}_{i,j}(z,s) - \lambda_{i,j}( (x+sv,v))| \leq \epsilon \bar{M}(z),
\]
where $z=(x,v)$; 
 and (B) that for any future time time $n\epsilon<T$ for positive integer $n$, that 
\[
\mbox{E}_z[\bar{M}(\bar{Z}_{n\epsilon})] \leq M(n\epsilon,z) <\infty,
\]
where $\bar{Z}_{t}$ is that state of the approximate PDMP at time $t$ and expectation is with respect to this state assuming $\bar{Z}_0=z$.

Part (A) follows (i) as the two rates are identical at $s=0$, i.e.  $\bar{\lambda}_{i,j}(z,0)=\lambda^{i,j}((x,v))$ and (ii) both rates have bounded derivative with respect to $s$ (this is simple to show for $\bar{\lambda}_{i,j}(z,s)$ from its definition, and is by assumption for $\lambda_{i,j}( (x+sv,v))$). The constant $\bar{M}(z)$ can be defined by to be the sum of the  maximum of each of these derivatives for $0\leq s \leq \epsilon_0$. Part (B) will then follow immediately as the region of possible values of $\bar{Z}_{n\epsilon}$ within a finite time interval is compact.

\section{Comparison with Stochastic Bouncy Particle Sampler}\label{app: comparison SBPS}
In this section, we compare SG-PDMPs with the Stochastic Bouncy Particle Sampler (SBPS), and its adaptive preconditioned version (pSBPS) as presented in  \citet{pakman2017stochastic}. The SBPS and pSBPS algorithms attempt to tackle a similar underlying problem as our paper: how to approximately implement a PDMP sampler with sub-sampling. The challenge lies in how to efficiently sample events of the PDMP with sub-sampling. As we shall explain below, our approach is simpler and computationally more efficient than SBPS and pSBP.

There are a number of substantive and practical differences between this approach and our algorithms. 
Our approach discretises time, and simulates events with a constant rate within each time-interval. This constant rate is simple to calculate, and involves accessing a single data point and calculating the current event rate associated with the process. 
By comparison, SBPS involves fitting a linear model to the rate based on the observed rates for mini-batches of data for rejected events since the last event. The SBPS approach needs to account for the randomness across the choice of mini-batch, and deal with the challenge of fitting a (probably incorrect) linear model to few (initially just one) data points. Together this means that SBPS uses much larger mini-batch sizes: the authors considered mini-batch sizes of approximately $n = 0.1 N$ compared to $n=1$ for our Euler scheme (a lower mini-batch can affect negatively the performance  of SBPS as the thinning rate grows linearly with $N/n$ -- see Equation~10 in \citet{pakman2017stochastic}). The uncertainty in fitting the linear model means one has to take conservative upper bounds, which can also lead to a resulting loss of efficiency.

We consider two Baysian problems: a Bayesian linear regression and a Bayesian logistic regression. In both examples, we set an improper uniform prior. In both cases, we set the dimension to be $d = 10$ and we simulate $N = 10^5$ covariates $X_{i,j} \sim \cN(0, \mathrm{I}_d))$. $y$ was simulated from the model with the true parameters $x^\star_i \sim \cN(-5,\, 100)$ (for the linear regression, we re-scaled the noise with $\sqrt{N}$). For both problems, we perform analogous experiments and we set the sub-sample batch size for the SBPS to be equal to 10\%$N$ as in \citet{pakman2017stochastic} and  all the other parameters suggested therein. Note that the need for SBPS to scale minibatch size with $N$ means that the computational advantages of our approach will grow (linearly in $N$) for larger data sets.

For both problems, we compare SBPS and pSBPS with SG-PDMPs in two numerical experiments. To make the comparison fair, for all implementations we have thinned the continuous-time output so that, for the same CPU cost, each algorithm outputs the same number of samples. In the first experiment we initialise all algorithms far away from  the bulk of the posterior measure, by setting $x_0 = \cN(0, 100^2)$. 

With this initialisation, we observe that, in the logistic regression problem, SBPS (respectively pSBPS) fails to approximate the Bouncy Particle Sampler with sub-sampling: it estimates an upper bound which fails to bound the real rate more than $50\%$ (respectively $22\%$). 
This undesirable behaviour translates in an algorithm which does not converge to the posterior density. In contrast, SG-PDMPs converge to the bulk of the posterior measure, for different values of step-size $h$. In the linear regression example, SBPS and pSBPS converge to the bulk of the distribution, but the rate of convergence is significantly slower compared to SG-PDMPs.

\begin{figure}[!htb]
   \begin{minipage}{0.48\textwidth}
     \centering
     \includegraphics[width=.7\linewidth]{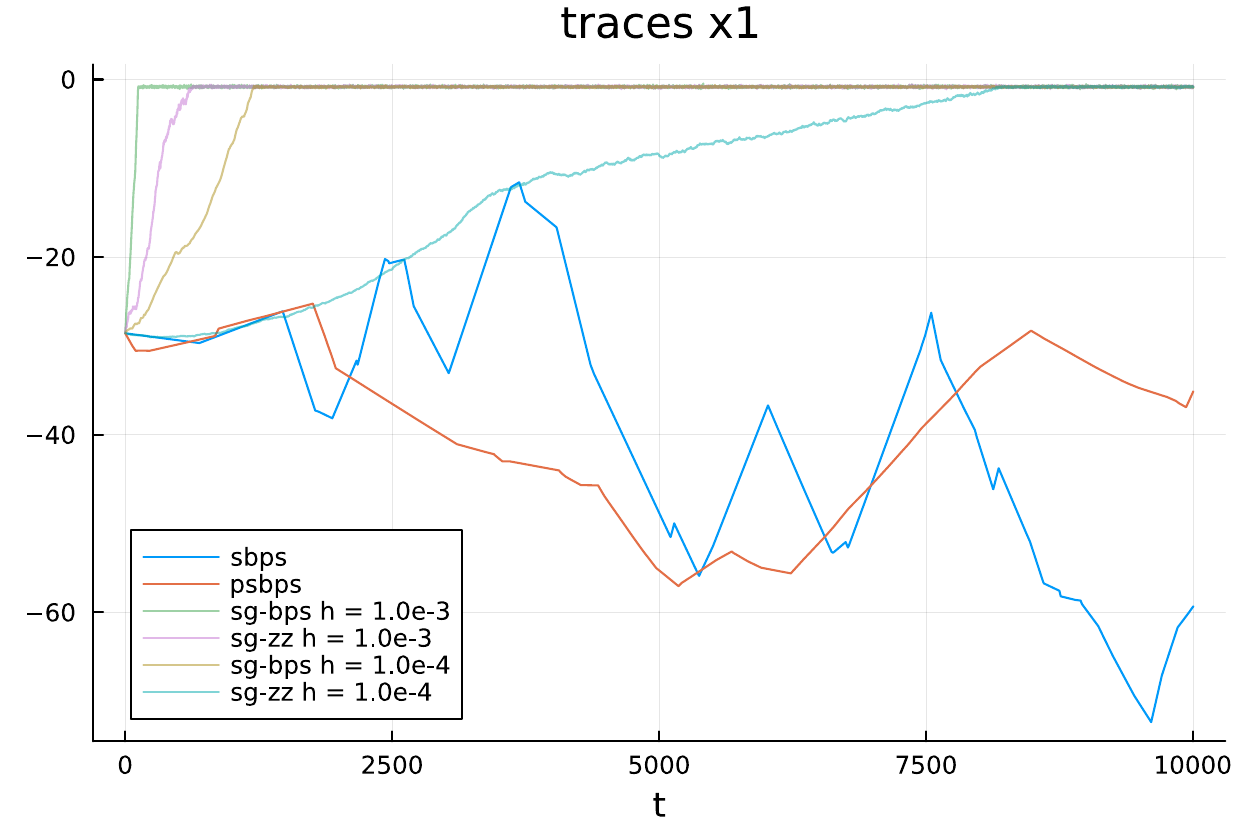}
     \caption{\label{fig: comparison SBPS1} Traces of the first coordinate of SBPS, SG-BPS, SG-ZZ (with step-sizes $10^{-3}, 10^{-4}$) relative to the \textbf{Bayesian logistic regression model}. All algorithms where initialised far away from the bulk of the posterior density (first experiment).}
   \end{minipage}\hfill
   \begin{minipage}{0.48\textwidth}
     \centering
     \includegraphics[width=.7\linewidth]{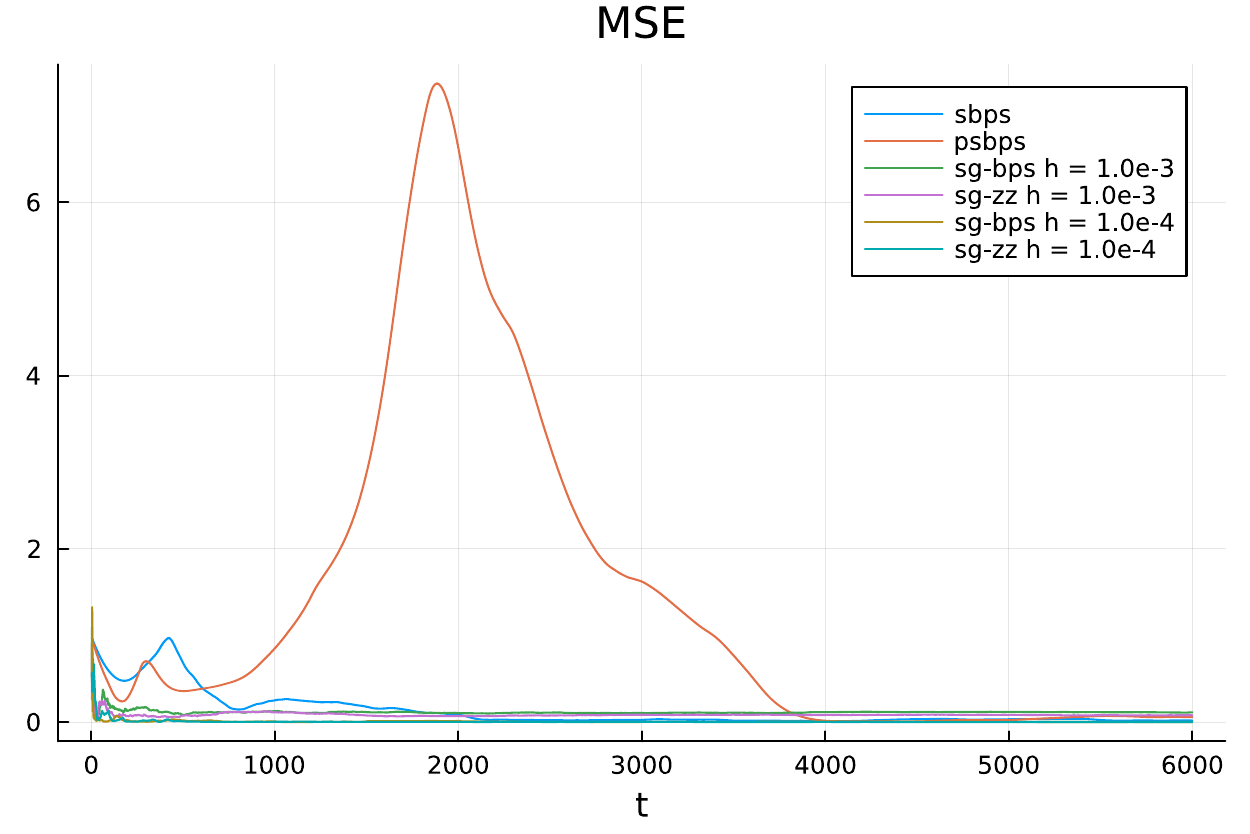}
     \caption{\label{fig: comparison SBPS2 bis} $\epsilon(t \div 2,t)$ for $t = 1,2,\dots,6000$ relative to the \textbf{Bayesian logistic regression model}. All algorithms where initialised near the mode of the posterior (second experiment).}
   \end{minipage}
\end{figure}

\begin{figure}[!htb]
   \begin{minipage}{0.48\textwidth}
     \centering
     \includegraphics[width=.7\linewidth]{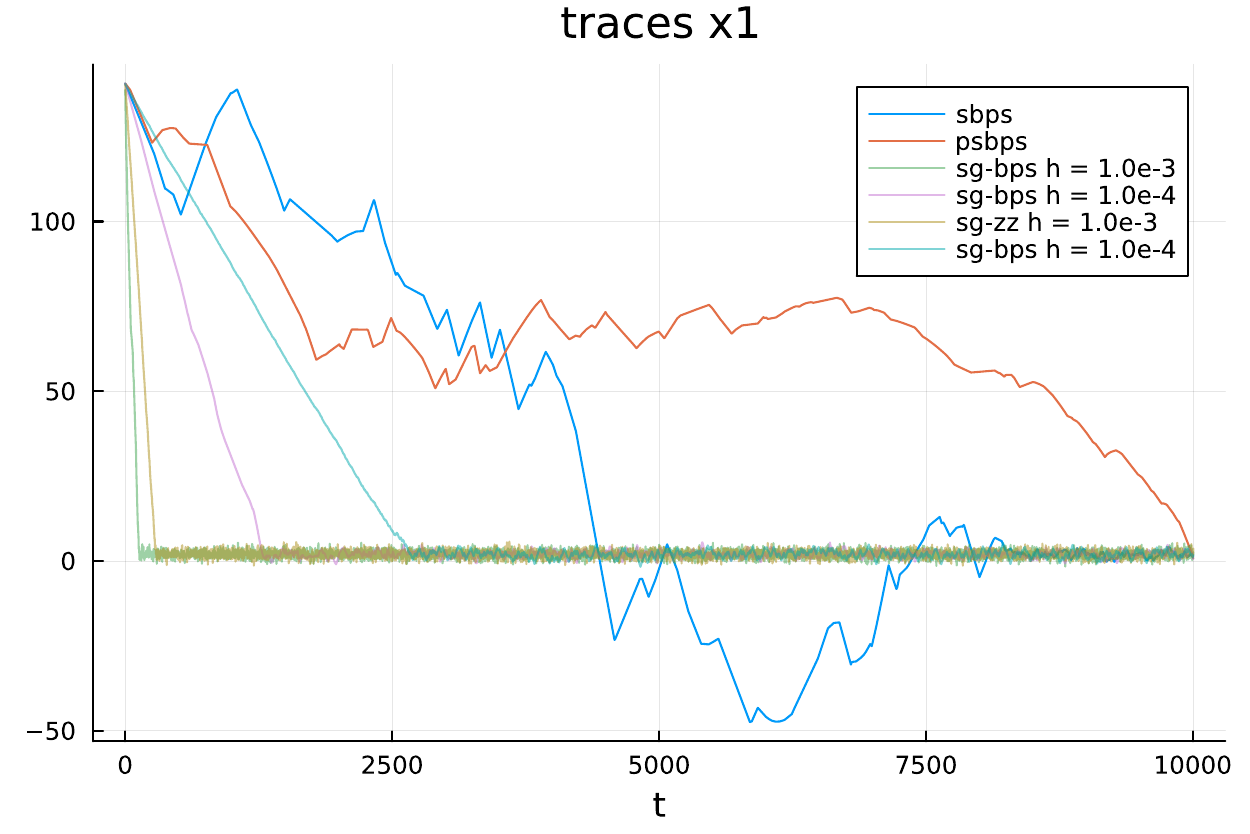}
     \caption{\label{fig: comparison SBPS5} Traces of the first coordinate of SBPS, SG-BPS, SG-ZZ (with step-sizes $10^{-3}, 10^{-4}$) relative to the \textbf{Bayesian linear regression model}. All algorithms where initialised far away from the bulk of the posterior density (first experiment).}
   \end{minipage}\hfill
   \begin{minipage}{0.48\textwidth}
     \centering
     \includegraphics[width=.7\linewidth]{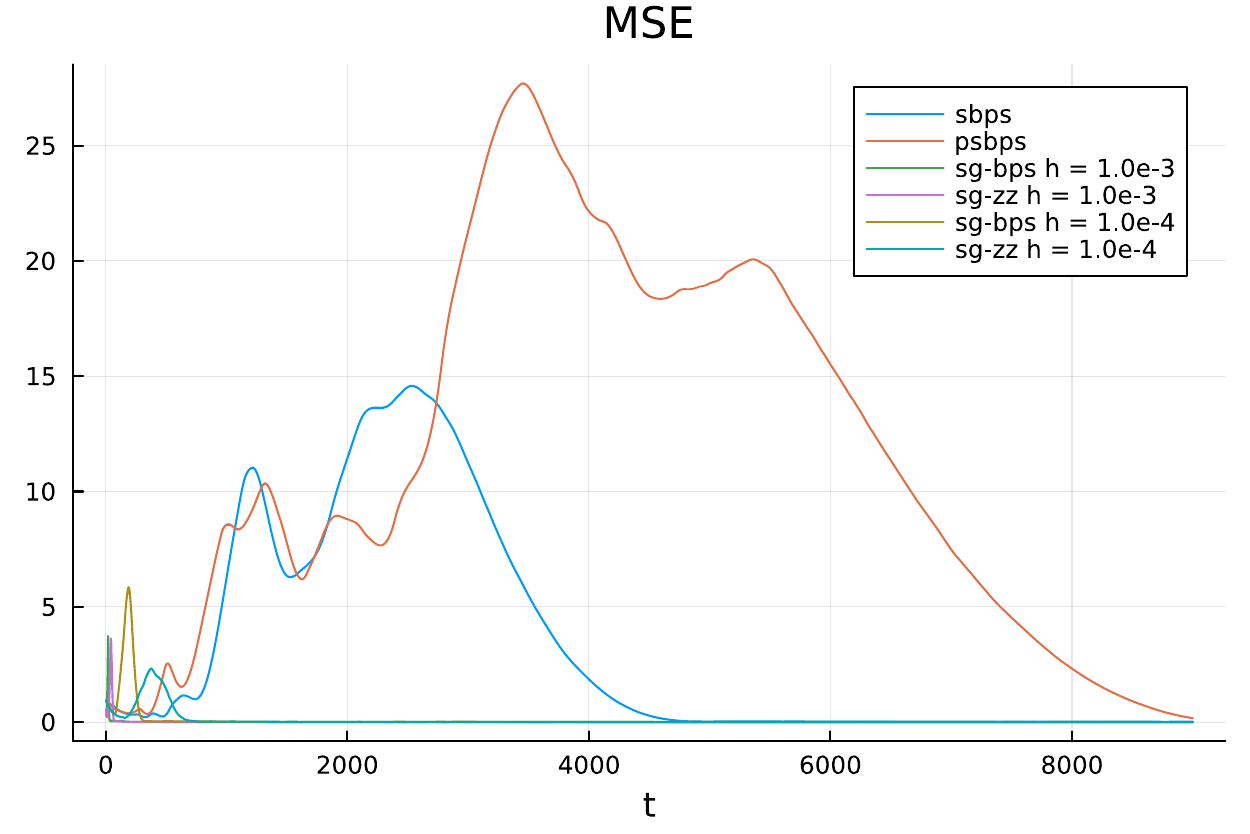}
     \caption{\label{fig: comparison SBPS4} $\epsilon(t \div 2,t)$ for $t = 1,2,\dots,6000$ relative to the \textbf{Bayesian linear regression model}. All algorithms where initialised near the mode of the posterior (second experiment).}
   \end{minipage}
\end{figure}

Figure~\ref{fig: comparison SBPS1} and Figure~\ref{fig: comparison SBPS5} display the trace of the first coordinate of SBPS, SG-BPS, SG-ZZ with step-size equal to $h = 10^{-3}$ and $h = 10^{-4}$ for the logistic regression and linear regression examples.

\begin{figure*}[!htbp]
\centering
\includegraphics[width=0.3\textwidth]{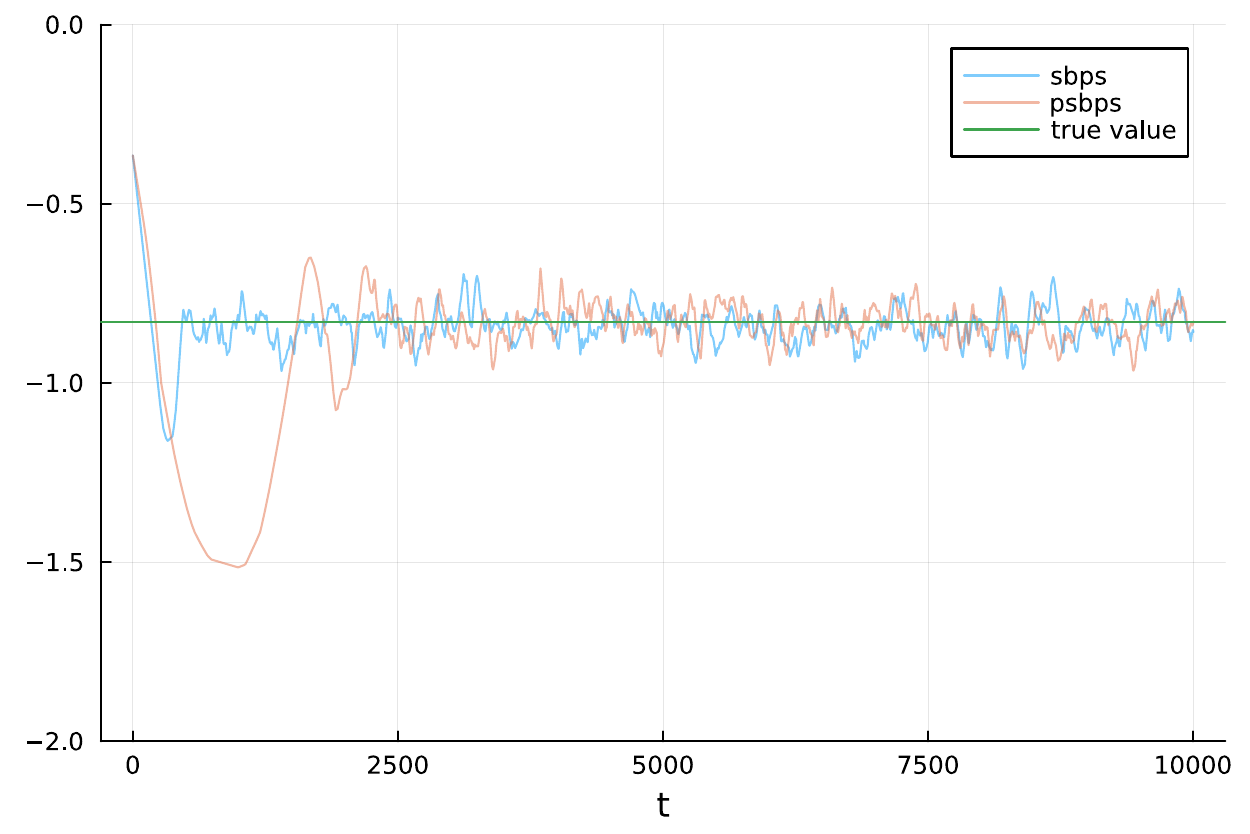} 
\includegraphics[width=0.3\textwidth]{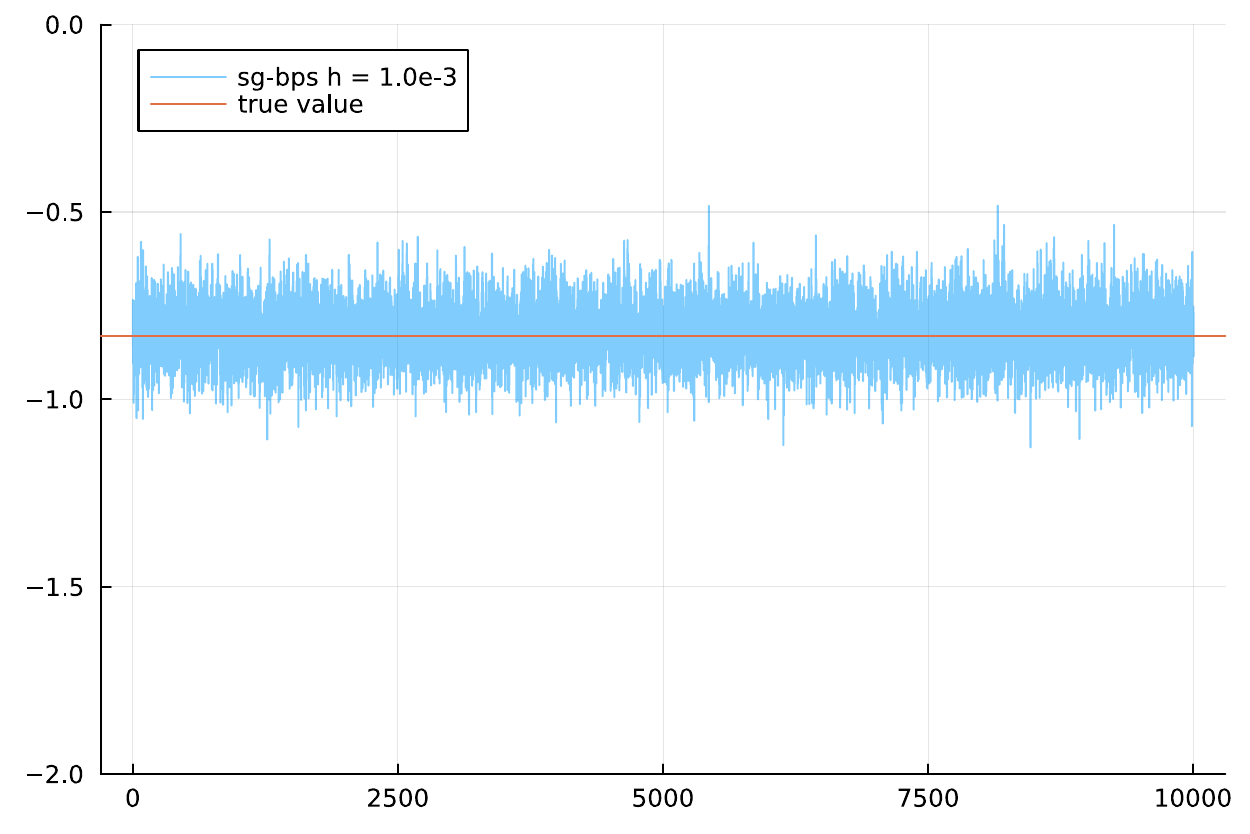} 
\includegraphics[width=0.3\textwidth]{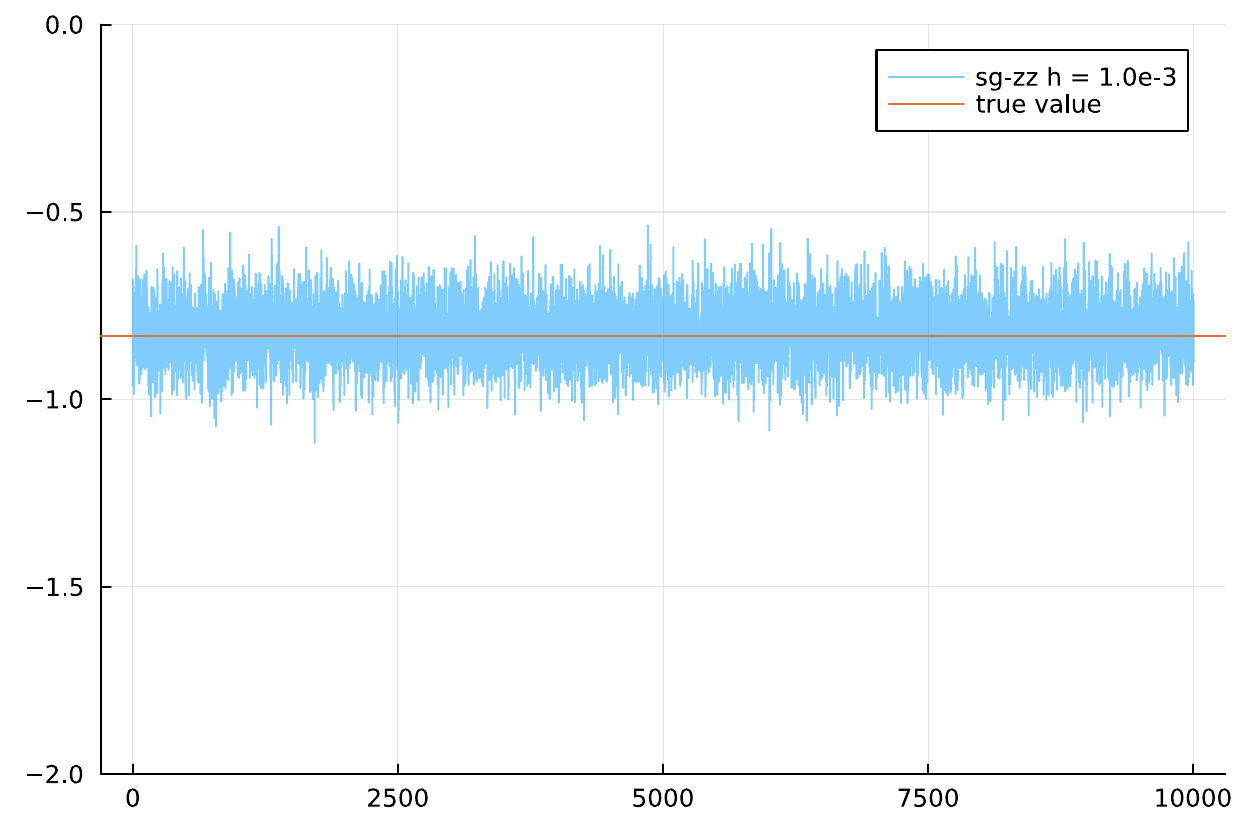} 
\includegraphics[width=0.3\textwidth]{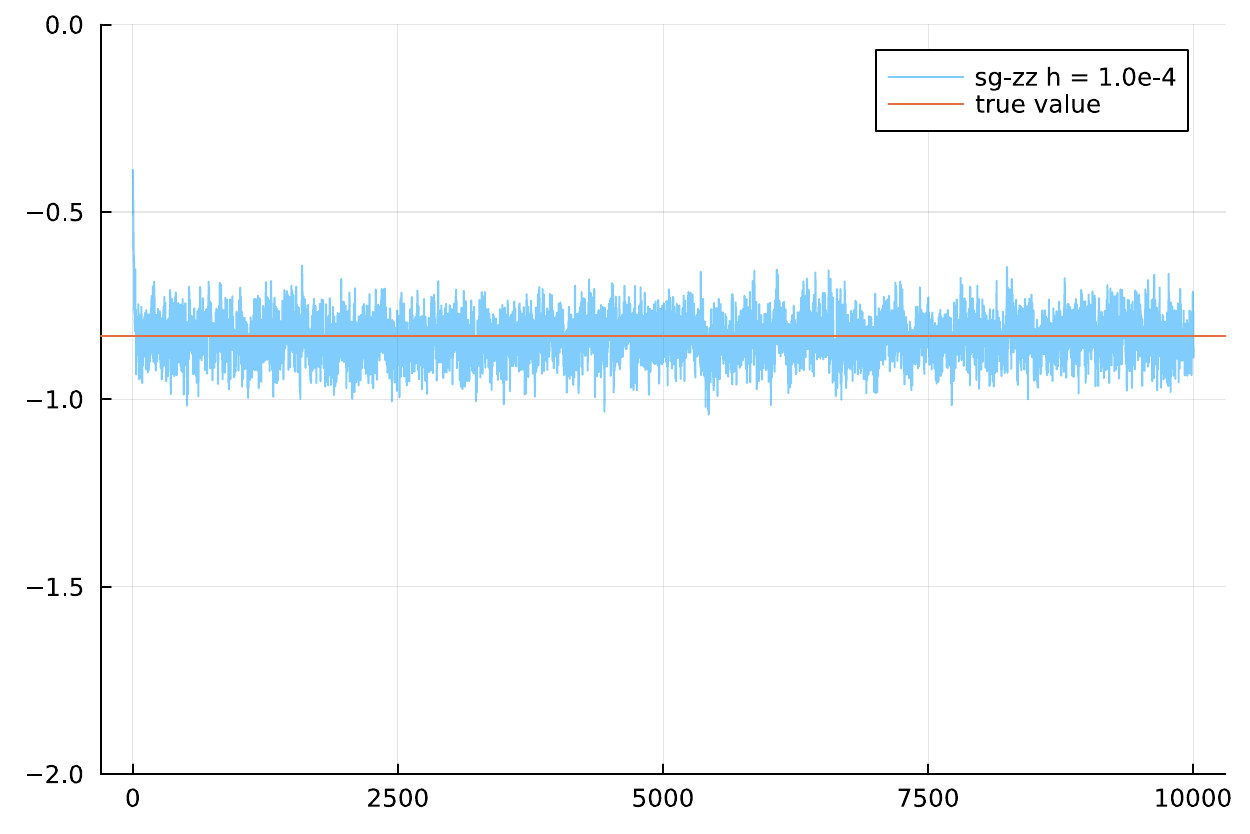} 
\includegraphics[width=0.3\textwidth]{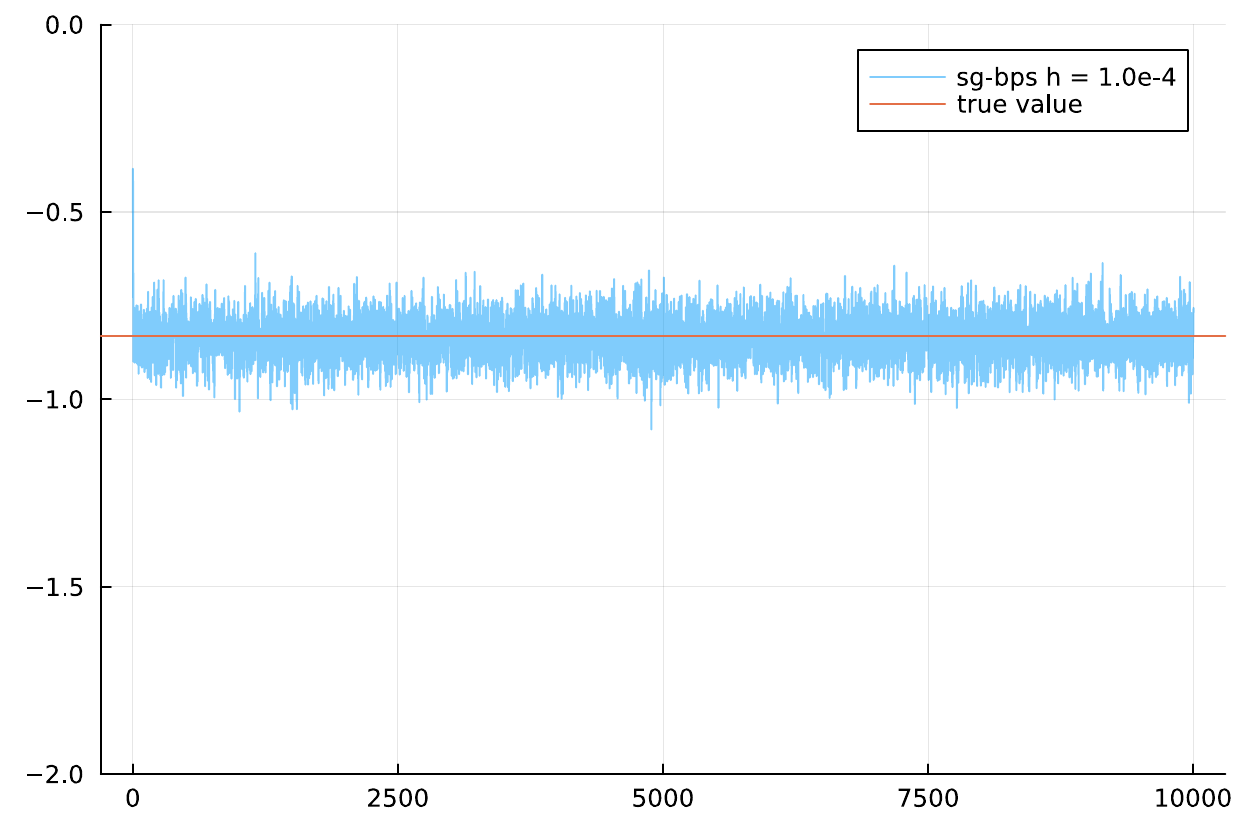} 
\includegraphics[width=0.3\textwidth]{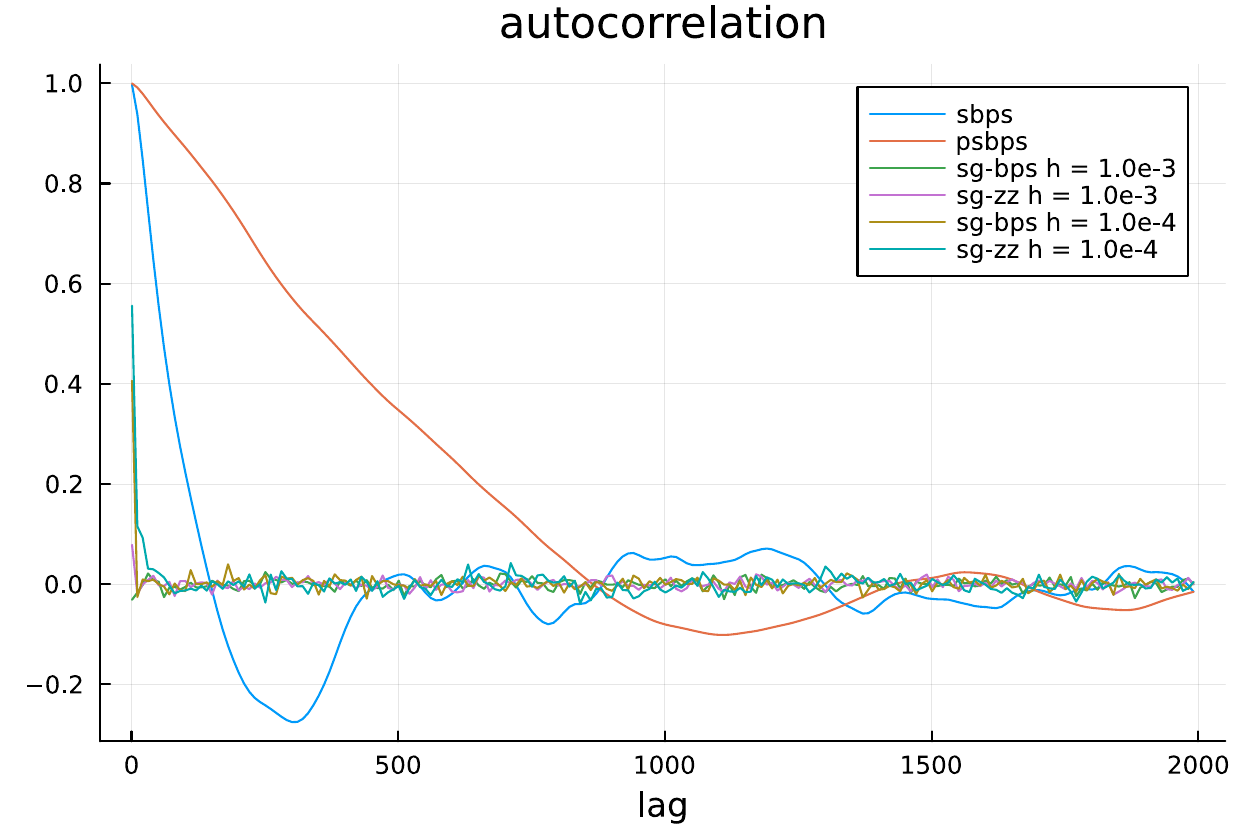} 
\caption{\label{fig: comparison SBPS2} From the top-left to the bottom-right. Traces  and auto-correlation function of the first coordinate of SBPS, SG-BPS, SG-ZZ (with step-sizes $10^{-3}, 10^{-4}$) relative to the \textbf{Bayesian logistic regression model}. All algorithms where initialised near the mode of the posterior (second experiment).}
\end{figure*}

\begin{figure*}[!htbp]
\centering
\includegraphics[width=0.3\textwidth]{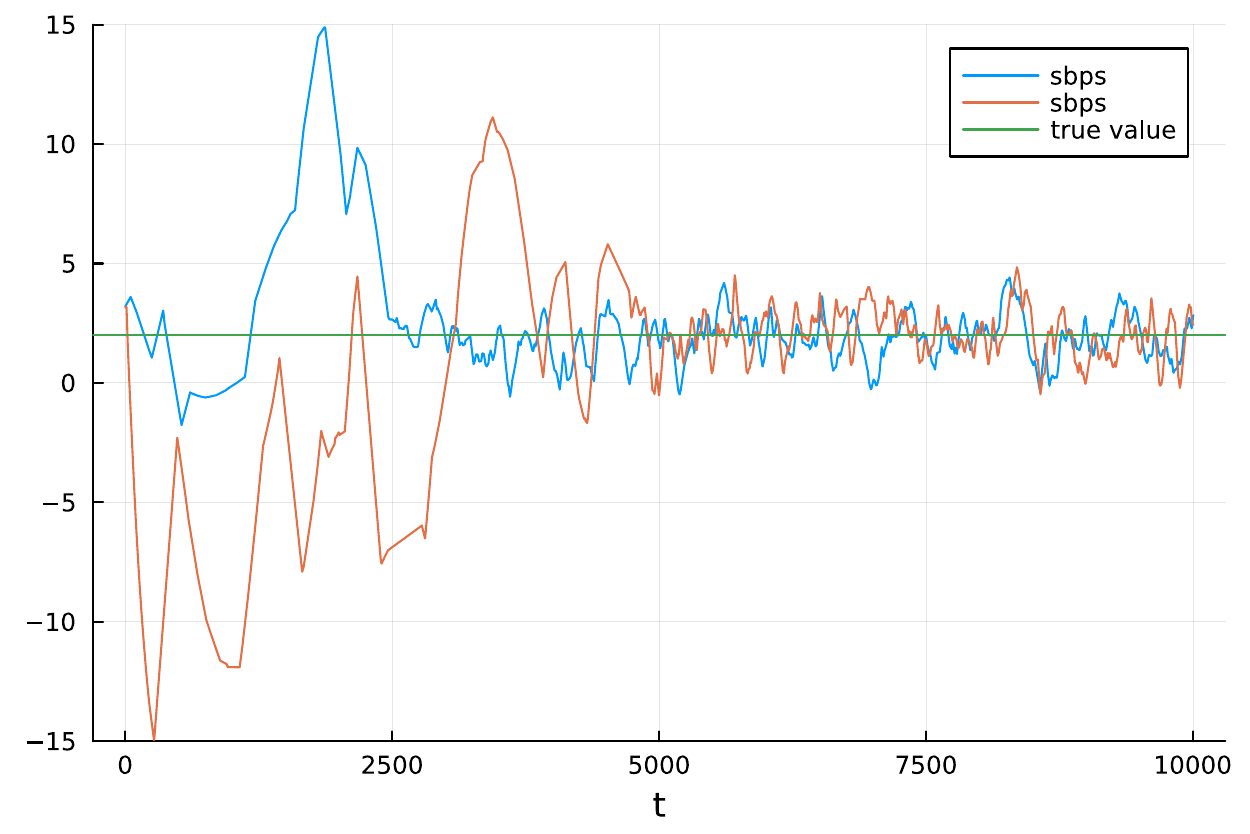} 
\includegraphics[width=0.3\textwidth]{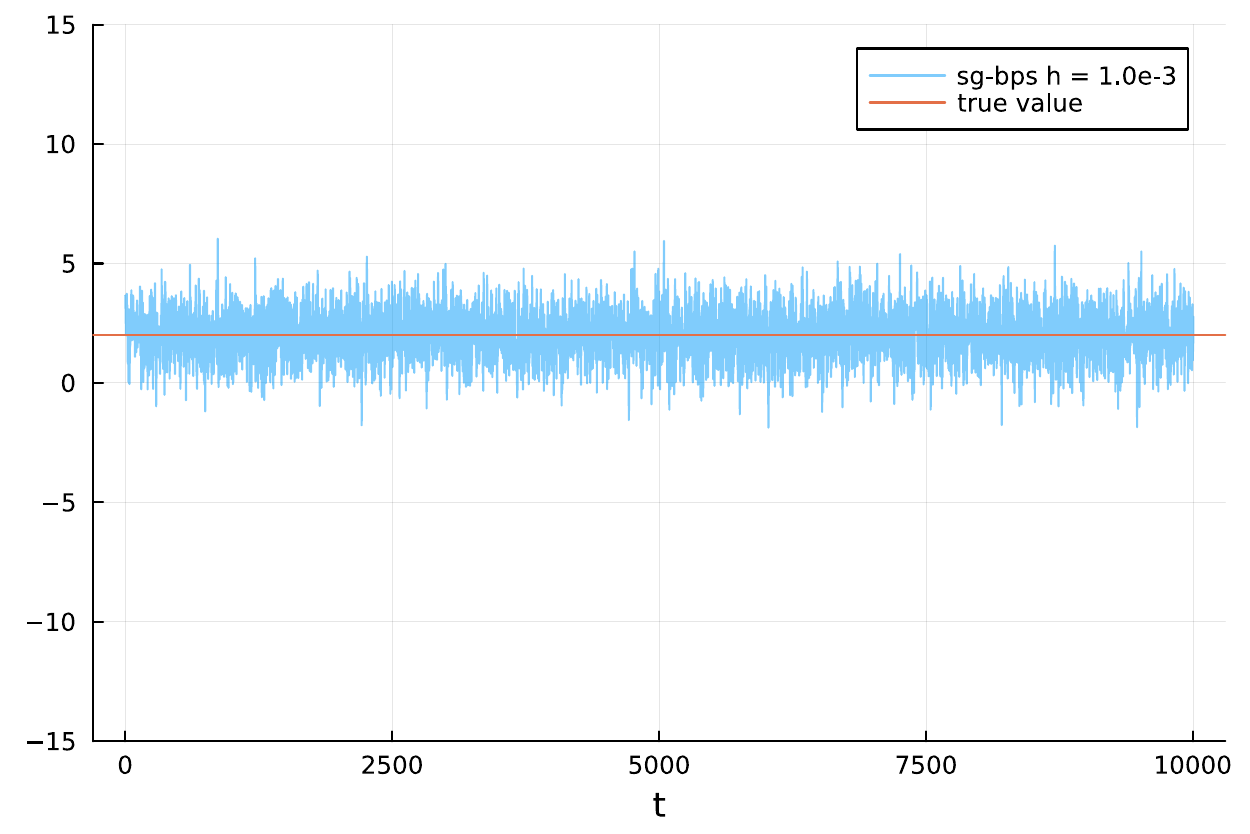} 
\includegraphics[width=0.3\textwidth]{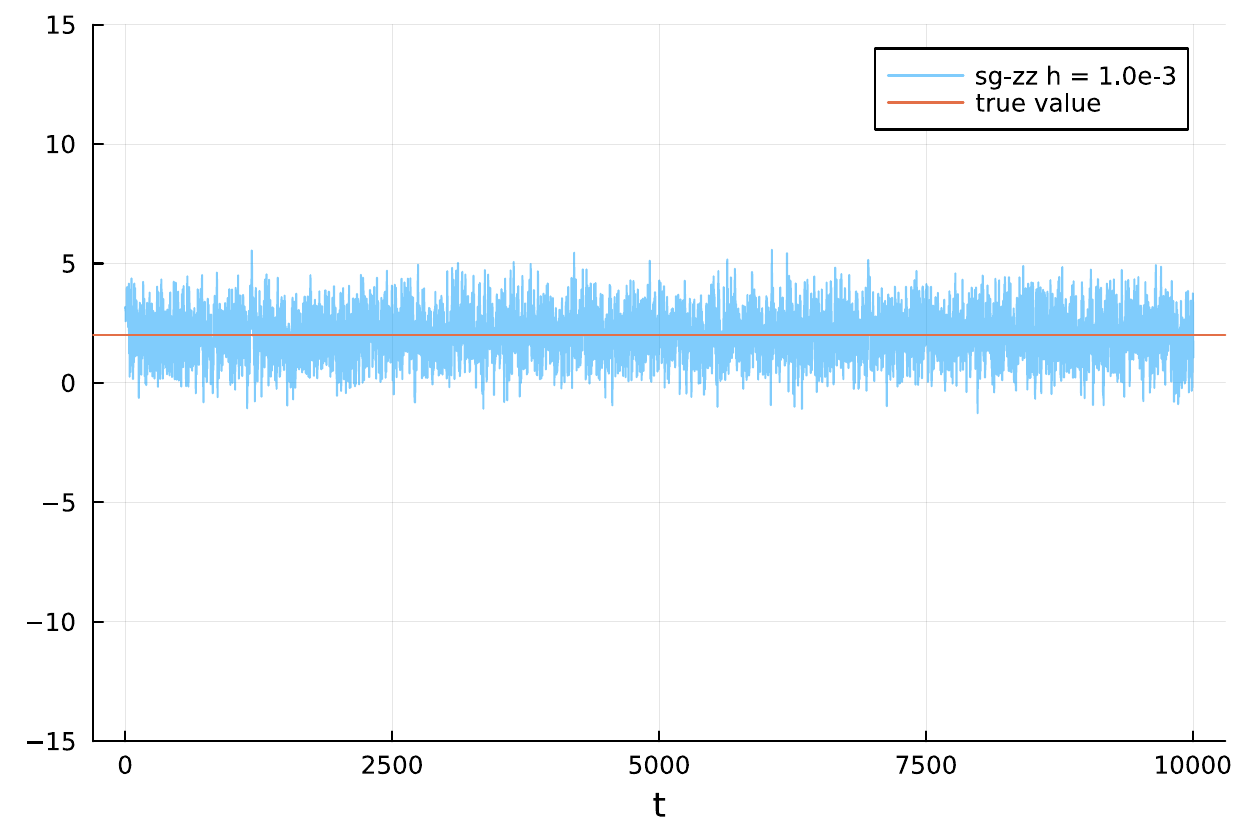} 
\includegraphics[width=0.3\textwidth]{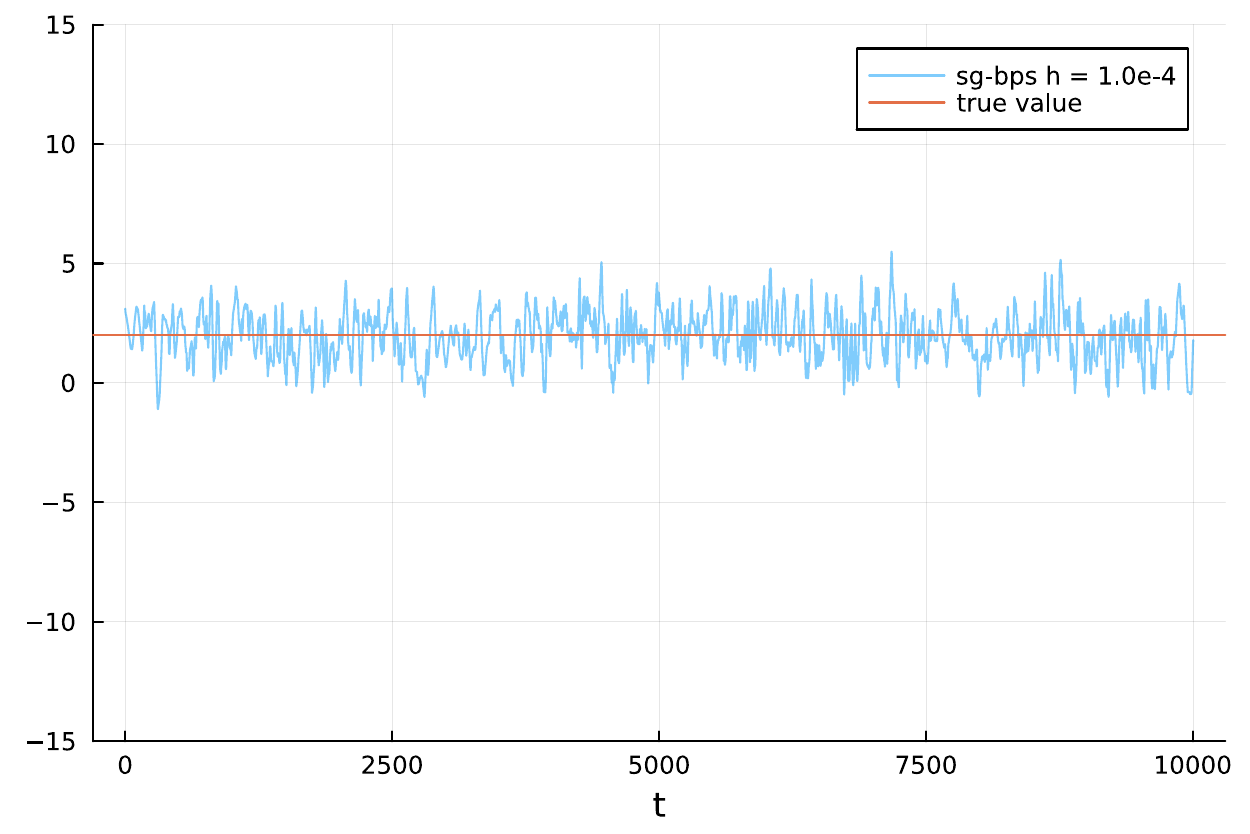} 
\includegraphics[width=0.3\textwidth]{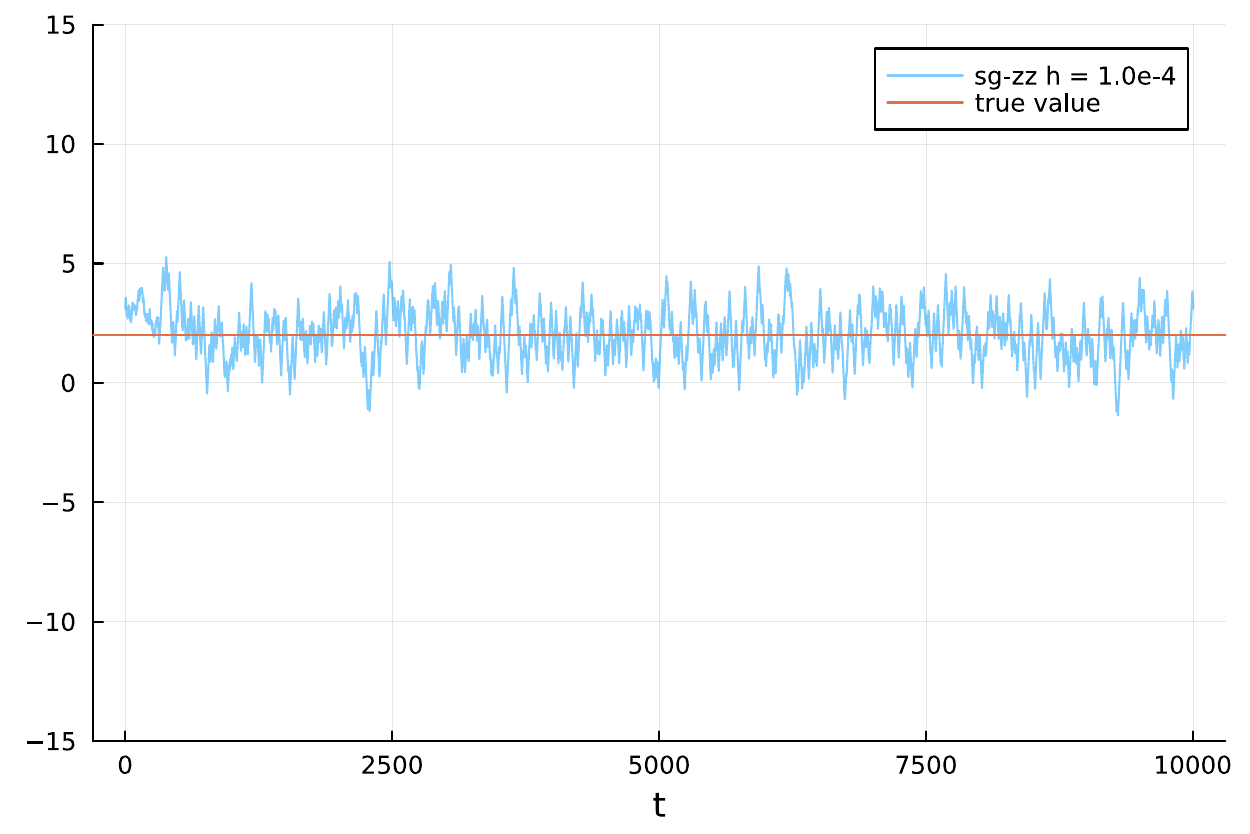} 
\includegraphics[width=0.3\textwidth]{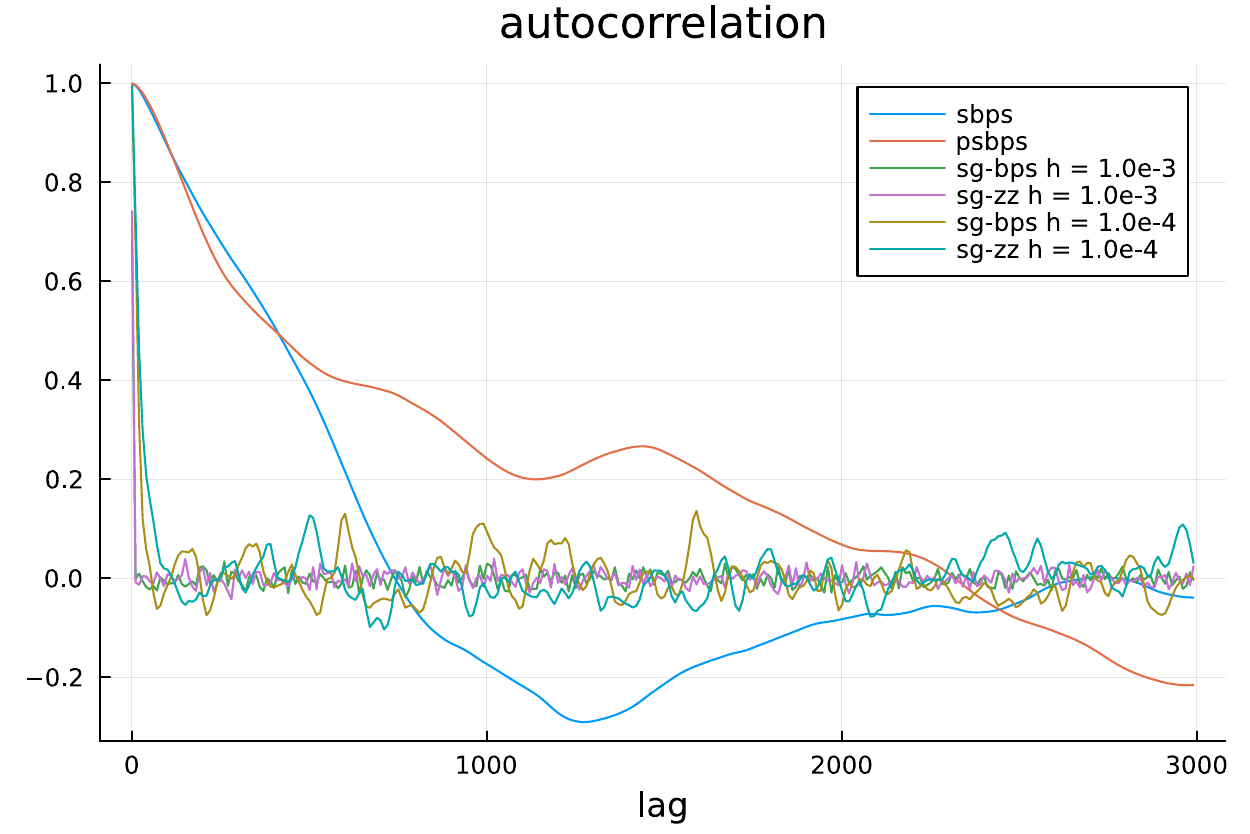} 
\caption{\label{fig: comparison SBPS3} From the top-left to the bottom-right. Traces  and auto-correlation function of the first coordinate of SBPS, SG-BPS, SG-ZZ (with step-sizes $10^{-3}, 10^{-4}$) relative to the \textbf{Bayesian linear regression model}. All algorithms where initialised near the mode of the posterior (second experiment).}
\end{figure*}

In the second experiment, we initialise all algorithms close to the posterior mode. In this setting, SBPS behaves better: for example, for the logistic regression, SBPS (respectively pSBPS) fails to upper bound the real rate $3\%$ ($4\%$) of the time. 
However, when comparing SBPS to SG-PDMPs, we notice that SBPS is substantially worse at mixing for both of the problems considered. We show in Figure~\ref{fig: comparison SBPS2}  and in Figure~\ref{fig: comparison SBPS3} the traces of the first coordinate fo SBPS, SG-BPS and SG-ZZ (with $h = 10^{-3},\,10^{-4}$) and the auto-correlation functions for the logistic and linear regression problem. Furthermore, for each algorithm we compute 
$$
\epsilon(t_1,t_2) = \frac{1}{d}\sum_{i=1}^d \left(\frac{\hat \sigma_{i}^{(t_1,t_2)} - \sigma_i}{\sigma_i}\right)^2
$$
where $\hat \sigma_{i}^{(t_1,t_2)}, \, t_1 \le t_2$ is the empirical standard deviation of the $ith$ coordinate computed with the output with burnin equal to $t_1$ and total number of sample points equal to $t_2$ and $\sigma_i$ is the empirical standard deviation computed with a long-run ($10^8$ iterations) of the SG-BPS with small step-size $h = 10^{-5}$. Figure~\ref{fig: comparison SBPS2 bis} and Figure~\ref{fig: comparison SBPS5} show for each algorithm $\epsilon(\lfloor\frac{t}{2}\rfloor,t)$ for different values of $t$ respectively for the logistic and linear regression problem.

\section{Stochastic Gradient ZZ and BPS} \label{sec:sgzzbps}

In Algorithms \ref{alg:sg-zz} and \ref{alg:sg-bps} we describe the specific implementation of stochastic gradient versions of the Zig-Zag sampler and the Bouncy Particle Sampler.

\begin{algorithm}[hbt!]
\caption{Stochastic gradient Zig-Zag (SG-ZZ) sampler}\label{alg:sg-zz}
\begin{algorithmic}
\REQUIRE $(x,v) \in E$, step size $\epsilon > 0$, time horizon $T>0$. 
\STATE $t = 0$, $\delta_t  = \epsilon$
\WHILE{$t < T$}
\STATE $J \sim \mathrm{Unif}(1,2,\dots,N)$
\STATE $\tau_i \sim \mathrm{Exp}( 
\lambda^{i,J}(z)
),\mbox{ for } i=1,\ldots,d$ 

\STATE $\tau=\min_{i=1,\ldots,d}\,\tau_i, \, i^* =\argmin \tau_i$
\IF{$\tau < \delta_t$}
\STATE $\delta_t \gets \delta_t - \tau, \, x \gets x + v\tau, \,t \gets t + \tau$
\STATE $v \gets \mathrm{R}_{i^*}(v)$  \COMMENT{Flip $i^*$ velocity component}
\ELSE
    \STATE $x \gets x + v\delta_t, \,t \gets t + \delta_t$
    \STATE $\delta_t = \epsilon$
    \STATE Save $(x,t)$
\ENDIF
\ENDWHILE
\end{algorithmic}
\end{algorithm}

\begin{algorithm}[hbt!]
\caption{Stochastic gradient Bouncy Particle Sampler (SG-BPS)}\label{alg:sg-bps}
\begin{algorithmic}
\REQUIRE $(x,v) \in E$, step size $\epsilon > 0$, time horizon $T>0$. 
\STATE $t = 0$, $\delta_t  = \epsilon$
\WHILE{$t < T$}
\STATE $J \sim \mathrm{Unif}(1,2,\dots,d)$
\STATE $\tau \sim \mathrm{Exp}( 
\lambda^{J}(z)
)$
\IF{$\tau < \delta_t$}
\STATE $v^{(\mathrm{new})} = F^J_x(v)$ \COMMENT{Velocity reflection}
\STATE $\delta_t \gets \delta_t - \tau, \, x \gets x + v\tau, \,t \gets t + \tau$
\STATE $v = v^{\mathrm{(new)}}$
\ELSE
    \STATE $x \gets x + v\delta_t, \,t \gets t + \delta_t$
    \STATE $\delta_t = \epsilon$
    \STATE Save $(x,t)$
\ENDIF
\ENDWHILE
\end{algorithmic}
\end{algorithm}
\section{Comparison of SG-PDMPs with varying mini-batch size}\label{sec: Comparison of SG-PDMPs different batch-sizes}
\begin{figure}[h!]
    \centering
    \includegraphics[width = 0.95\textwidth]{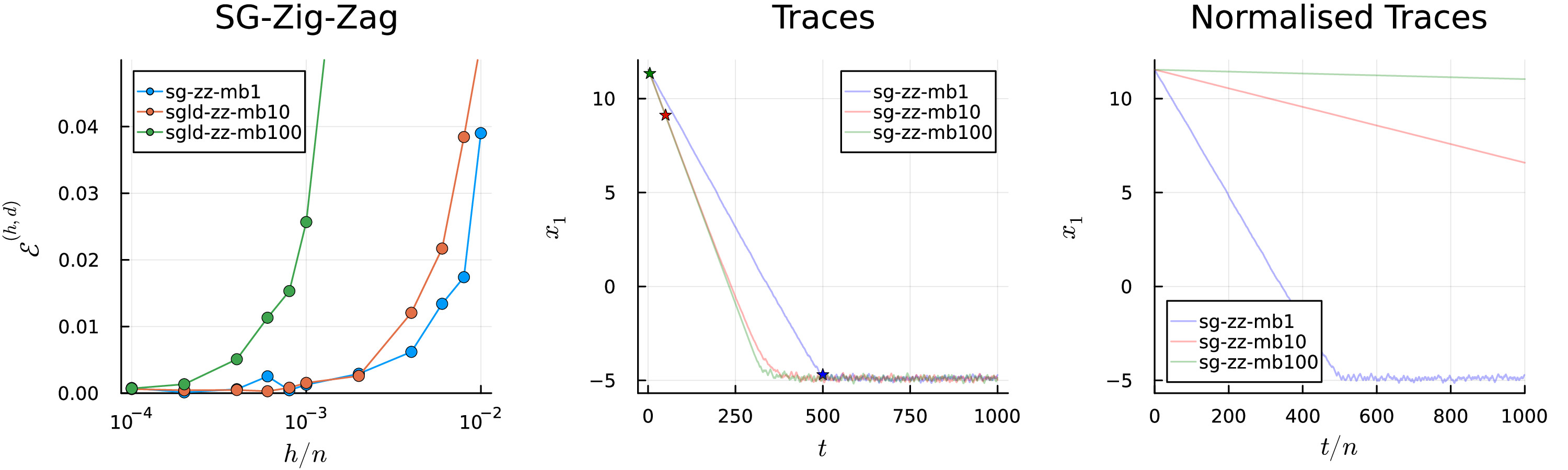}
    \caption{Left panel: error in the standard deviation estimation of each coordinate $\mathcal{E}^{(h,d)}$ (y-axis) against the step-size divided by the mini-batch size $\frac{h}{n}$ (x-axis, on a log-scale). We run all algorithms for the same time horizon so that $\frac{h}{n}$ is a proxy for the inverse of the running time of each algorithm. Middle panel: traces of one coordinate relative of SG-ZZ with step-size h = $10^{-3}$ and mini-batch size equal to 1 (blue), 10 (red), $10^2$ (green). All algorithms were initialised out of stationarity. The star symbols indicate the location of each Markov chain after $25\times10^3$ gradient evaluations. Right panel: traces normalised by the mini-batch size.}
    \label{fig: varying minibatch sizes}
\end{figure}

PDMP samplers with subsampling and SG-PDMPs, as presented in Section~\ref{subsec: pdmps with subsampling}-\ref{sec: stochastic gradient PDMP samplers}, can be extended by using, at every iteration, a batch-size of $n < N$ random data points instead of a single one. We view the ability to use a mini-batch size of 1 to be a key advantage of our method and, in most contexts, we would expect worse performance (relative to CPU cost) for larger mini-batches. In this section we support our claim with numerical experiments. A summary of our experiments is shown in Figure~\ref{fig: varying minibatch sizes}
 where we run the SG-Zig-Zag algorithm with different mini-batch sizes for a linear regression problem with $10^6$ observations. As can be seen, for a fixed CPU cost using a mini-batch size of 1 is the best – this is most strikingly shown in the right-hand plot.

\section{Sticky Zig-Zag sampler} \label{sec:szz}
In what follows, we present the stochastic gradient sticky Zig-Zag (SG-SZZ) sampler which approximates the sticky Zig-Zag (\citet{bierkens2021sticky}). In this case, the target distribution  is assumed to be of the form
\begin{equation}
    \label{eq: sticky target}
    \pi(\dd x) \propto \exp(-U(x))\prod_{i=1}^d (\dd x_i + \frac{1}{\kappa_i} \delta_0(\dd x_i) 
\end{equation}
where $ U(x) = \sum_{j=1}^N U_j(x)$. With simple algebra, it is not difficult to see that the taregt measure given by a smooth log-likelihood $\ell(x) = \sum_{i=1}^N \ell_i(x)$ and spike and slab prior 
\begin{equation}
    \label{eq: supp spike and slab prior}
\pi_0(\dd x) = \otimes_{i=1}^d  (1-w_i)\pi_i(x_i) \dd x_i + w_i\delta_0(\dd x_i)
\end{equation}
is of the form of equation~\eqref{eq: sticky target}, with 
\[
U_i(x) = C - \ell_i(x) - \frac{1}{N}\sum_{j=1}^d \log(\pi_j(x_j)), \qquad \kappa_i =\frac{1-w_i}{w}\pi_i(0),
\]
for some constant $C$ indepent from $x$.
In algorithm~\ref{alg:sg-pdmp_sticky}, we present the SG-SZZ as a minor modification of the SG-ZZ, enabling to approximate measures which are of the form of ~\eqref{eq: sticky target}.
\begin{algorithm}[hbt!]
\caption{Stochastic gradient sticky Zig-Zag (SG-SZZ) samplers}\label{alg:sg-pdmp_sticky}
\begin{algorithmic}
\REQUIRE $(x,v, s) \in E$, step size $\epsilon > 0$, time horizon $T>0$,  sticky parameter $\kappa \in (0, \infty)^{d}$. 
\STATE $t = 0$, $\delta_t = \epsilon$
\STATE $v^{c} = v$ \COMMENT{Copy of velocity}
\STATE $A = \{1,2,\dots,d\}$, $A^c = \emptyset$
\WHILE{$t < T$}
\STATE $J \sim \mathrm{Unif}(1,2,\dots,d)$
\STATE $\tau = \min_{i \in A}\{\tau_i \sim \mathrm{Exp}(
\lambda^{i,J}(z)
)\}$ 
\STATE $\tau^\star = \min_{i \in A}\{ t_i^\star \colon x_i + v_i t_i = 0\}$
\STATE $\tau^\circ = \min_{i \in A^c}\{t_i^\circ \sim \mathrm{Exp}(k_i)\}$
\IF{$\tau < \min(\delta_t, \tau^\star, \tau^\circ)$}
\STATE $x = x + v\tau, \, t = t + \tau, \, \delta_t \gets \delta_t - \tau$
\STATE $v \gets \mathrm{R}_{j}(v)$ where $j :=  \mathrm{argmin}_{i \in A}\{ \tau_i\}$
\ELSIF{$\tau^\star < \min(\delta_t,\tau^\circ)$}
    \STATE $x = x + v\tau^\star, \, t = t + \tau^\star,  \, \delta_t \gets \delta_t - \tau^\star$
    \STATE $A^c \gets j^\star $ where $j^\star:=  \mathrm{argmin}_{i \in A}\{ \tau_i^\star\}$
    \STATE $v_i = 0$ \COMMENT{Stick }
\ELSIF{$\tau^\circ< \delta_t$}
    \STATE $x = x + v\tau^\circ, \, t = t + \tau^\circ, \, \delta_t \gets \delta_t - \tau^\circ$
    \STATE $A \gets j^\circ$ where $j^\circ :=  \mathrm{argmin}_{i \in A^c}\{\tau_i^\circ\}$ 
    \STATE $v_i = v_i^{c}$ \COMMENT{Unstick}
\ELSE
    \STATE $x = x + v\delta_t, \, t = t + \delta_t$
    \STATE $\delta_t = \epsilon$
    \STATE Save $(t, x)$
\ENDIF

\ENDWHILE
\end{algorithmic}
\end{algorithm}
\section{Logistic regression}\label{app: logistic}
\subsection{Stein discrepancy kernel}\label{app: Stein discrepancy kernel}
 The Stein discrepancy kernel evaluated on $K$  sample points with dimenision $d$ is defined as
\[
\sum_{k=1}^d \sqrt{\sum_{i,j = 1}^K \frac{\kappa_k(x^{(i)}, x^{(j)})}{K^2}}
\]
where
\[
\kappa_k(y,z) = \partial_{x_k}U(x)\partial_{y_k} U(y) \kappa(x,y) + \partial_{x_k} U(x) \partial_{y_k} \kappa(x,y) + \partial_{y_k} U(y)\partial_{x_k} \kappa(x,y) + \partial_{x_k}\partial_{y_k}\kappa(x,y).
\]
Following \citet{nemeth2021stochastic} Section 4, we choose $\kappa(x,y) = (c^2 + \|x-y\|^2)^{\beta},\, \beta \in (-1,0), \, c>0$.  \citet{gorham2017measuring} shows that this metric is able to detect both poor mixing and bias from the target distribution.
\subsection{Further simulations}
\begin{figure}[h!]
    \centering
    \includegraphics[width = 0.49\textwidth]{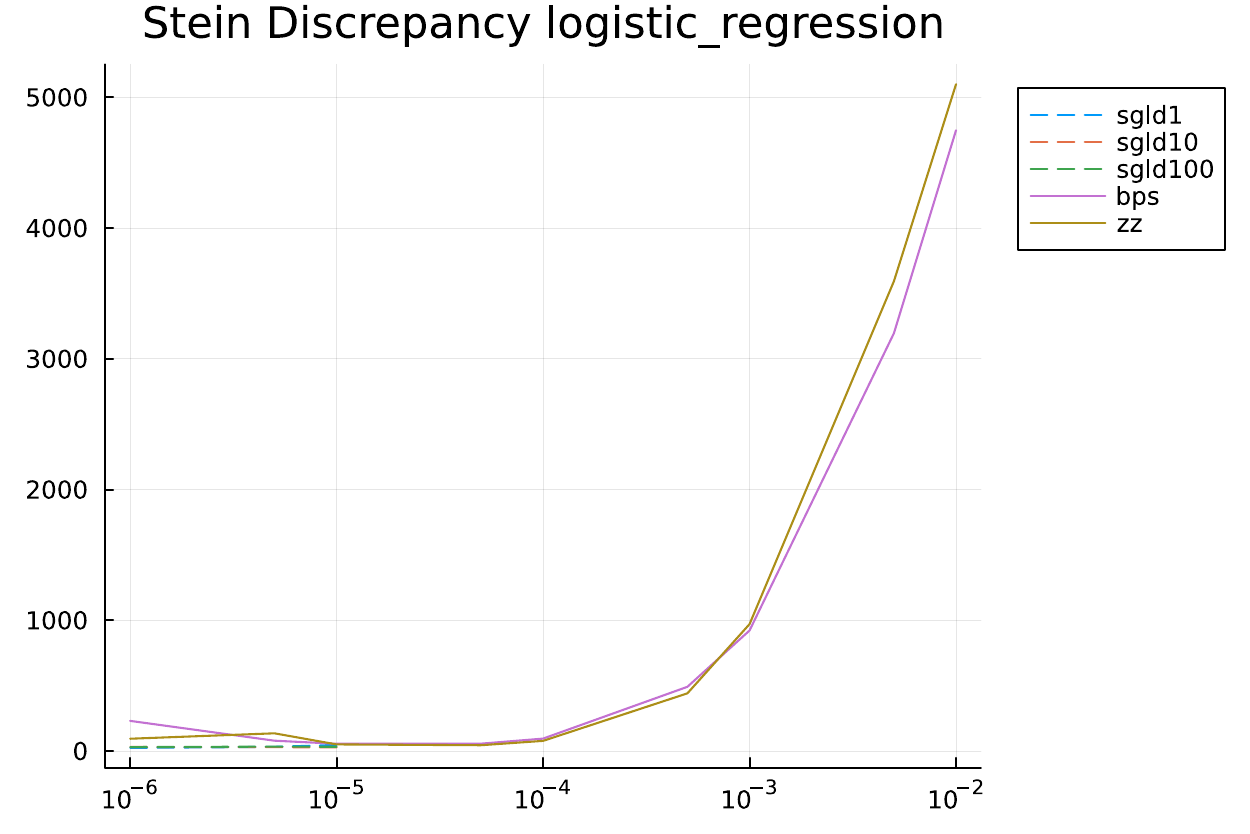}
    \includegraphics[width = 0.49\textwidth]{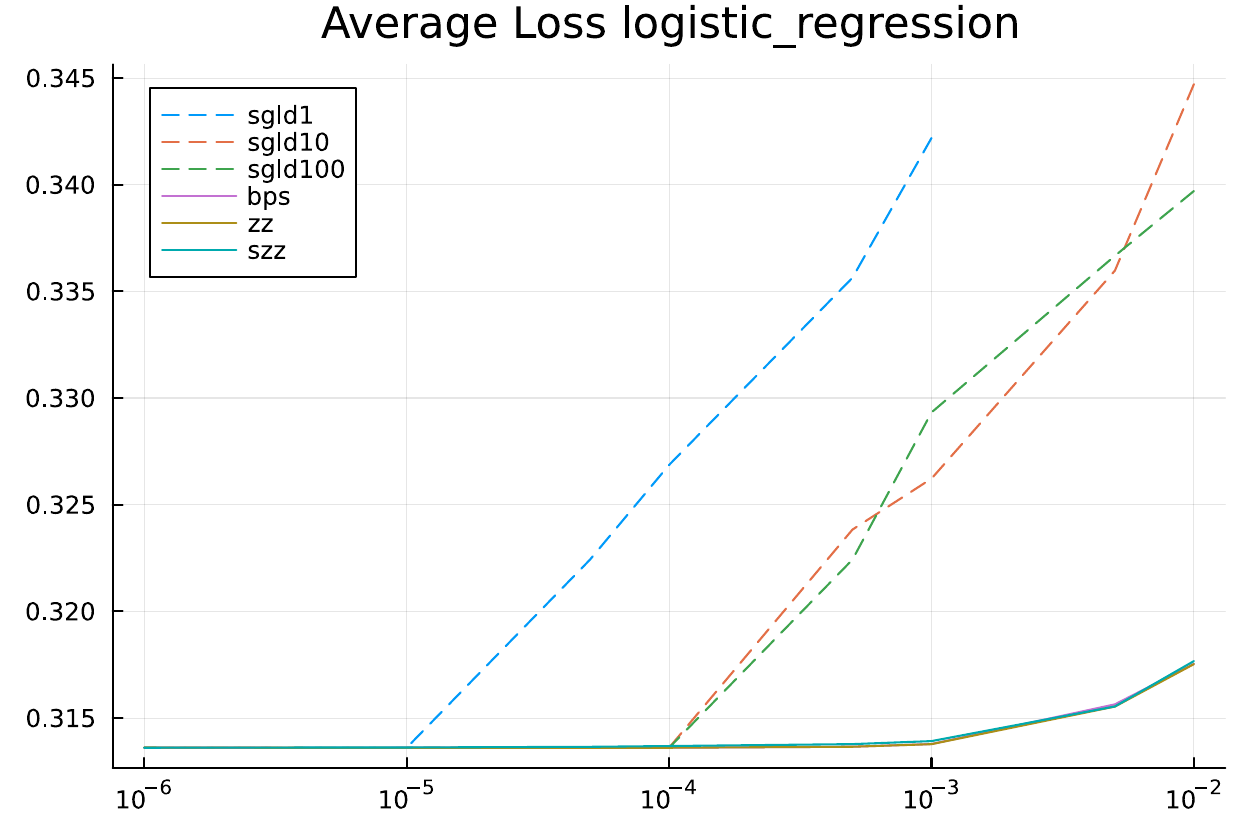}
    \caption{Stein discrepancy kernel (left panel) and average loss function on $M = 10$ random permutation of train and test sets (right panel) for the logistic regression as a function of the stepsize $h$ ($x$-axis in log-scale). Dashed lines corresponds to SGLD algorithms, solid lines to SG-PDMPs. For SGLD, the Stein Discrepancy kernel was finite only for $h \le 10^{-5}$.}
    \label{fig: stein and prediction}
\end{figure}

We compute the Stein discrepancy  kernel relative to the traces obtained with SGLD with 1, 10, 100 minibatch sample points and SG-ZZ and SG-BPS. We run each algorithm varying $h$ and we plot in Figure~\ref{fig: stein and prediction}, top panel the results.
Figure~\ref{fig: stein and prediction}, bottom panel, displays the loss function of the samplers above and additionaly SG-SZZ with spike-and-slab prior as in Equation~\eqref{eq: supp spike and slab prior} with $w_i = 0.5, \, i=1,2,\dots,d$. The loss function has been averaged over $M = 10$ random permutation of train and test datasets, as described in Section~4.1 of the main manuscript. 
For this model, we set the loss function of Equation~\eqref{eq: loss function} equal to
\[
\ell(X_i,y_i, x) = y_i \log\left(\frac{1}{1 + \exp(- \langle x, X_i \rangle)}\right) + (1-y_i) \log \left(1- \frac{1}{1 + \exp(- \langle x, X_i \rangle)}\right). 
\]

\section{Bayesian neural networks}\label{app: Bayesian neural networks}
\begin{figure}[h!]
    \centering
    \includegraphics[width = 0.49\textwidth]{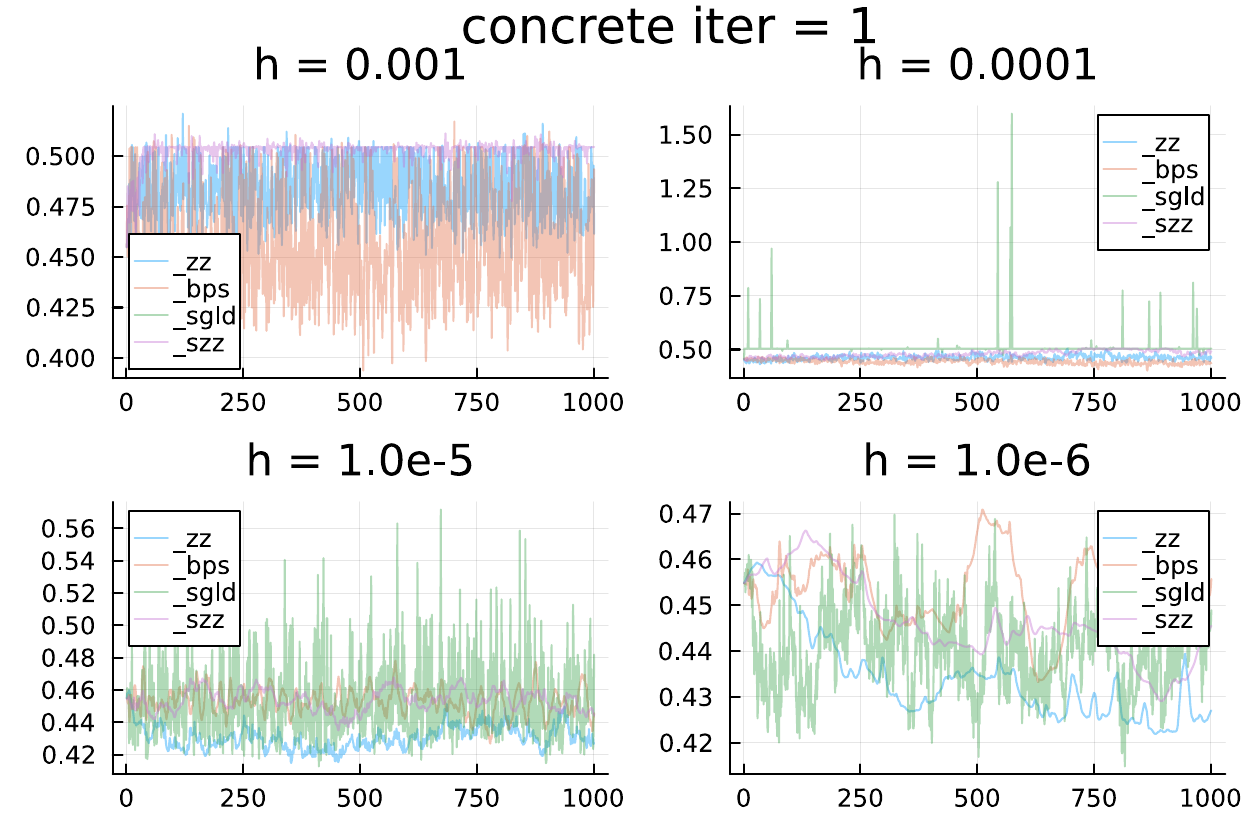}
    \includegraphics[width = 0.49\textwidth]{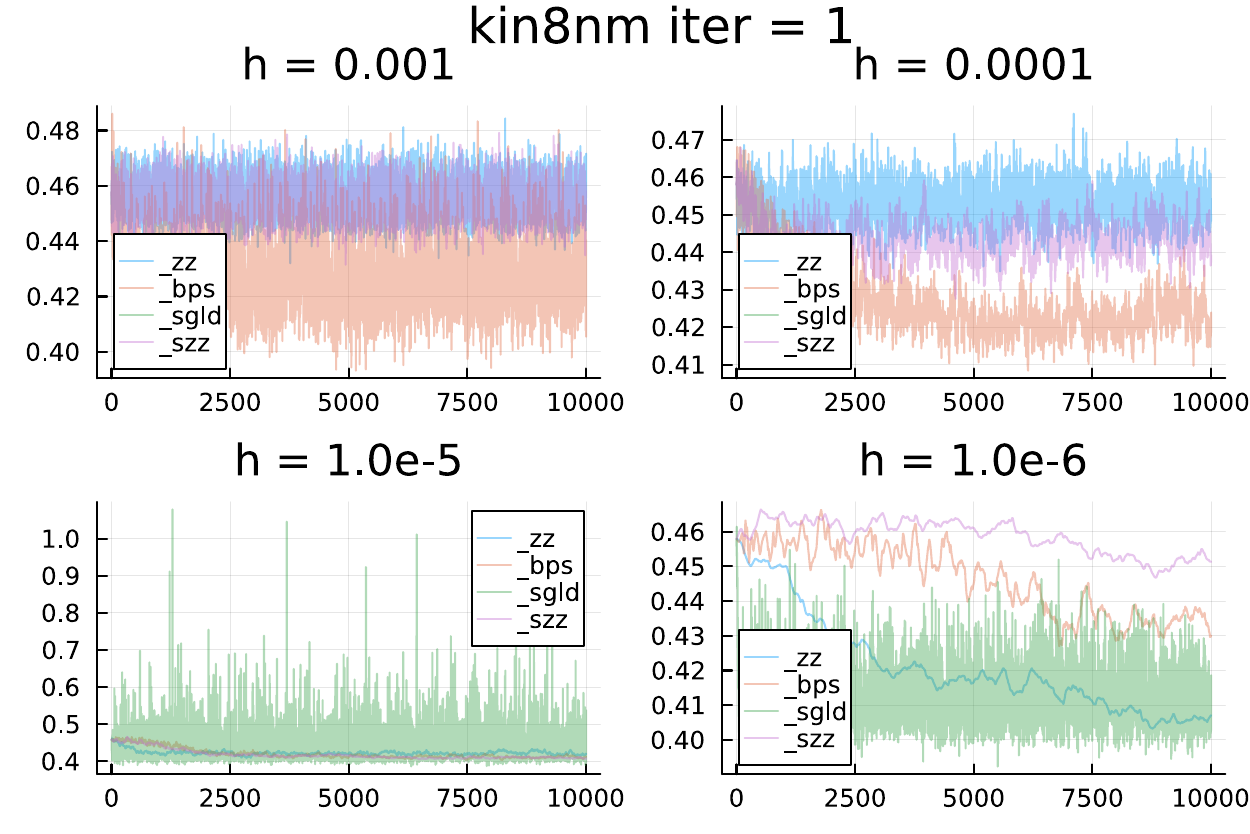}
    \caption{Trace of the loss function for each sampler for different step-sizes $h$ relative to the first permutation of the training and test set of the dataset `concrete' (left panel) and `kin8nm' (right panel). SGLD was unstable for $h = 10^{-3}$ for the dataset `concrete' and for $h = 10^{-3}, 10^{-4}$ for the dataset `kin8nm'.}
    \label{fig:loss nn supplement}
\end{figure}

In this section, we show the trace of the loss function (Figure~\ref{fig:loss nn supplement}) and report the average loss on the test sets relative to the traces obtained by running SGLD, SG-ZZ, SG-BPS, SG-SZZ on $M = 3$ different permuation of train and test sets for the other 2 datasets from \emph{UCI machine learning repository}\footnote{https://archive.ics.uci.edu/} labeled as `concrete' and `kin8nm' (Table~\ref{tab: loss nn supplement}) which have respectively $N = 1031, N = 45731$ and $p = 8, p = 9$ covariates. 

\begin{table}[h!]
\begin{center}
\begin{tabular}{l|llll}\\
 Concrete & \multicolumn{4}{c}{\textbf{h}}\\
  \textbf{sampler}      & $10^{-3}$ & $10^{-4}$ & $10^{-5}$ & $10^{-6}$ \\
        \hline
SGLD    &   \emph{NaN}         & \emph{0.5219} & 0.4625 & \textbf{0.4513} \\
SG-ZZ   &     0.5103 & 0.4803 & \textbf{0.4526} & 0.4739 \\
SG-BPS  &   \textbf{0.4681} & \textbf{0.4606} & 0.4759 & \emph{0.4885} \\
SG-SZZ  &   0.5167  &0.5017 & \emph{0.4932}  & 0.4877 \\  
\end{tabular}
\begin{tabular}{l|llll}\\
 Kin8mn & \multicolumn{4}{c}{\textbf{h}}\\
  \textbf{sampler}      & $10^{-3}$ & $10^{-4}$ & $10^{-5}$ & $10^{-6}$ \\
        \hline
SGLD    &    \emph{NaN}        &  \emph{NaN}        & \emph{0.4480} & \textbf{0.4143} \\
SG-ZZ   &   0.4536 &   0.4465 & \textbf{0.4224} & 0.4250\\
SG-BPS  &    \textbf{0.4257}   & \textbf{0.4272} & 0.4230  & 0.4460\\
SG-SZZ  &      0.4520  &  0.4375 & 0.4233  & 0.4506\\
\end{tabular}
\end{center}
\caption{Average loss for each sampler for different values of step size $h$ relative to the dataset `concrete' (left table) and `kin8mn' (right table). In bold the best performance and italic the worst performance given a step size $h$.} \label{tab: loss nn supplement}
\end{table}


\end{document}